\documentclass[journal]{IEEEtran}
\usepackage{cite}
  
\ifCLASSINFOpdf
\else
\fi

\usepackage{graphicx}
\usepackage[caption=false,font=footnotesize]{subfig}
\usepackage{multirow}
\usepackage{ctable}
\usepackage{enumitem}
\usepackage{balance}

\newcommand{\changed}[1]{{#1}} 

\begin{document}

\title{Data, Depth, and Design: Learning Reliable Models \\for Skin Lesion Analysis}

\author{
    Eduardo Valle$^\ast$, Michel Fornaciali, Afonso Menegola, Julia Tavares,\\ Fl\'avia Vasques Bittencourt, Lin Tzy Li, Sandra Avila

    \thanks{
        E. Valle, M. Fornaciali, A. Menegola, and J. Tavares are affiliated to the Department of Computer Engineering and Industrial Automation (DCA) of the School of Electrical and Computing Engineering (FEEC), University of Campinas (UNICAMP), Campinas, Brazil.
        
  		E. Valle,  M. Fornaciali, A. Menegola, J. Tavares, L. T. Li, and S. Avila are affiliated to the RECOD Lab.
         
        F. Bittencourt is affiliated to the School of Medicine, Federal University of Minas Gerais (UFMG), Belo Horizonte, Brazil.
        
        L. T. Li is also affiliated to the Samsung Research and Development Institute Brazil, Campinas, Brazil.
        
        S. Avila is affiliated to the Institute of Computing (IC), UNICAMP.     
        
        $^\ast$Contact author: dovalle@dca.fee.unicamp.br, mail@eduardovalle.com.
    }
}

%

\maketitle

\begin{abstract}
Deep learning fostered a leap ahead in automated skin lesion analysis in the last two years. Those models, however, are expensive to train and difficult to parameterize.
\textit{Objective:} We investigate methodological issues for designing and evaluating deep learning models for skin lesion analysis. We explore ten choices faced by researchers: use of transfer learning, model architecture, train dataset, image resolution, type of data augmentation, input normalization, use of segmentation, duration of training, additional use of \textcolor{black}{Support Vector Machines}, and test data augmentation.
\textit{Methods:} We perform two full factorial experiments, for five different test datasets, resulting in 2560 exhaustive trials in our main experiment, and 1280 trials in our assessment of transfer learning. We analyze both with multi-way \textcolor{black}{analyses of variance (ANOVA)}. We use the exhaustive trials to simulate sequential decisions and ensembles, with and without the use of privileged information from the test set.
\textit{Results — main experiment:} Amount of train data has disproportionate influence, explaining almost half the variation in performance. Of the other factors, test data augmentation and input resolution are the most influential. Deeper models, when combined, with extra data, also help. \textit{--- transfer experiment:} Transfer learning is critical, its absence brings huge performance penalties. \textit{--- simulations:} Ensembles of models are the best option to provide reliable results with limited resources, without using privileged information and sacrificing methodological rigor.
\textit{Conclusions and Significance:} Advancing research on automated skin lesion analysis requires curating larger public datasets. Indirect use of privileged information from the test set to design the models is a subtle, but frequent methodological mistake that leads to overoptimistic results. Ensembles of models are a cost-effective alternative to the expensive full-factorial and to the unstable sequential designs.
\end{abstract}

\begin{IEEEkeywords}
Skin lesion analysis, deep learning, experimental design, model parameterization, cross dataset.
\end{IEEEkeywords}

%
\IEEEpeerreviewmaketitle

%
%
\section{Introduction}\label{sec:intro}

\textcolor{black}{While deep learning is well-established as the gold standard for image classification, it poses challenges to adapt for new tasks, due to a profusion of hyperparameters. Automated skin lesion analysis is no exception, with a state of the art that has sharply improved since the adoption of those models, but which still lacks reliable recommendations for their design.}

\textcolor{black}{Recent works on skin lesion classifiers present novel deep-learning models, each representing a given choice of hyperparameters. Those works' primary goal is improving the classification metrics on a given skin lesion dataset. Each paper compares a handful of models (representing different hyperparameterizations), concluding that some choice --- for example, the use of a given deep-learning architecture --- is better than the alternatives. Different authors recommend different hyperparameters (Section~\ref{sec:sota}), an apparent contradiction that arises because each work evaluates a limited number of factors/parameters spanning a small number of models.} 

\textcolor{black}{What literature currently lacks --- and this work intends to provide --- is a {\em systematic} evaluation of the hyperparameters of deep learning for skin lesion analysis. We propose an exhaustive evaluation of ten factors faced when designing deep-learning models.} We explore the variations of those factors in two full factorial designs. Our main experiment evaluates nine factors, for five different test datasets, resulting in $2^9\times5=2560$ treatments. Our assessment of transfer learning evaluates eight factors for five datasets, with $2^8\times5=1280$ treatments. We analyze those results using multi-way \textcolor{black}{analyses of variance (ANOVA)}, to obtain decisions that generalize across datasets.

\textcolor{black}{Our work is unprecedented in existing art, both in the scope of the experiments, and the rigor of the statistical analyses. Because the full factorial design used in our ANOVAs is too expensive for most practical contexts, we also simulate two alternative designs: the traditional sequential optimization (a single factor at a time) and ensembles of models. We show that the latter provides the best performance, at reasonable costs.}

This paper follows our participation at the International Skin Imaging Collaboration --- ISIC 2017 Challenge~\cite{codella2017skin}. At the time, with a strategy of aggressively optimizing our models, we were ranked first in melanoma classification~\cite{menegola2017recod}, with an \textcolor{black}{area under the receiver operating characteristic curve (AUC)} of 0.874. \textcolor{black}{Here we have different goals, focusing on understanding which factors of the deep learning approach most contributed for boosting performances.}

\textcolor{black}{The contributions of the paper are}:
\begin{itemize}
\item A main evaluation of nine factors affecting the design of deep learning for skin lesion analysis, together with their cross interactions up to the third level;
\item A quantitative appraisal of the importance of transfer learning, in perspective with seven of the nine factors analyzed in the main experiment;
\item \textcolor{black}{A demonstration of the ability of simple ensemble solutions to provide good results, without sacrificing methodological rigor};
\item \textcolor{black}{A critique of hyperparameter optimization, demonstrating that the customary procedure of optimizing hyperparameters and evaluating techniques on the same dataset leads to overoptimistic results};
\item State-of-the-art AUCs in all three classification tasks of the ISIC Challenge 2017/Part~3. \textcolor{black}{While the focus of this paper is not on increasing metrics; those results showcase both the power of ensembles, and the ability to obtain state of the art results while following strict methodological constraints};
\item Finally, our entire source code is available, and our procedures are explained step-by-step, from the acquisition of the data, until the generation of the tables and graphs of this paper\footnote{https://github.com/learningtitans/data-depth-design}. We are committed to make this work as reproducible as technically feasible.
\end{itemize}

We organized the remaining text as follows: we discuss the recent state of the art in Section~\ref{sec:sota}, and very briefly overview our participation on ISIC Challenge 2017. We detail our goals, materials, and methods in Section~\ref{sec:methods}. Experimental results and analyses appear in Section~\ref{sec:experiments}. Finally, we review and discuss the main findings in Section~\ref{sec:conclusion}.

%
%
\section{Survey of recent techniques} \label{sec:sota}

\textcolor{black}{In this Section, we survey existing art on automated skin lesion analysis. We focus on works published since 2016, when literature took a sharp turn towards deep learning. For relevant previous literature~\cite{celebi2007methodological,iyatomi2008improved}, we refer the interested reader to more comprehensive surveys~\cite{litjens2017survey,fornaciali2016towards}. We assume that the reader has passing familiarity with deep learning and its issues --- if needed, we refer the reader to a complete overview~\cite{LeCun2015deep}.}

\textcolor{black}{Recent works on skin lesion classifiers have the primary goal of improving classification metrics on a given skin lesion dataset. Because of the staggering number of design choices for deep learning, each author chooses a limited number of hyperparameters to evaluate, spanning a handful of prospective novel models. Usually, a paper is considered fit for publication when at least one of those prospective models shows better performance metrics than recent techniques.}

\textcolor{black}{In the next paragraphs, we discuss the hyperparameters raised most frequently by existing literature on deep-learning skin analysis. For each hyperparameter, we illustrate how different authors have taken different stances, indicating whether authors' opinions tend to converge or diverge on their choices.}

Existing works either train deep networks from scratch~\cite{sabbaghi2016deep,nasr2016melanoma,jia2017skin}; or reuse the weights from pre-trained networks ~\cite{yu2017automated,esteva2017dermatologist,menegola2017knowledge,lopez2017skin,harangi2017skin,codella2017deep}, in a scheme called \textbf{transfer learning}. Transfer learning is usually preferred, as it alleviates the main issue of deep learning for skin lesion analysis: way too small datasets --- most often comprising a few thousand samples. (Contrast that with the ImageNet dataset, employed to evaluate deep networks, with more than \textit{a million} samples.) Training from scratch is preferable only when attempting new architectures, or when avoiding external data due to legal/scientific issues. Menegola et al.~\cite{menegola2017knowledge} explain and evaluate transfer learning for skin lesion analysis in more detail. 

\textcolor{black}{In this paper, in our main experiment of hyperparameterization of deep learning, all treatments employ transfer learning. We also perform an experiment to quantify the importance of transfer learning, in which half the treatments employ transfer learning, and half do not.}

Whether using transfer or not, works vary widely in their choice of deep-learning architecture, from the relatively shallow (for today's standards) VGG~\cite{yu2017automated,menegola2017knowledge,lopez2017skin,ge2017exploiting}, mid-range GoogLeNet~\cite{yu2017automated,esteva2017dermatologist,harangi2017skin,yang2017novel,vasconcelos2017increasing}, until the deeper ResNet~\cite{menegola2017recod,harangi2017skin,codella2017deep,matsunaga2017image,bi2017automatic} or Inception~\cite{menegola2017recod,esteva2017dermatologist,devries2017skin}. On the one hand, more recent architectures tend to be deeper, and to yield better accuracies; on the other hand, they require more data and are more difficult to parameterize and train. Although high-level frameworks for deep learning have simplified training those networks, a good deal of craftsmanship is still involved. In this work, we contrast two architectures: \textbf{ResNet-101-v2}~\cite{He2016eccv}, and \textbf{Inception-v4}~\cite{szegedy2016inceptionv4}. Both are very deep and show outstanding performance on ImageNet, but Inception-v4 is considerably deeper and bigger than ResNet-101-v2.

Data augmentation is another technique used to bypass the need for data, while also enhancing networks' invariance properties. Augmentation creates a myriad of new samples by applying random distortions (e.g., rotations, crops, resizes, color changes) to existing samples. Augmentation provides best performance when applied to both train and test samples, but only the most recent skin lesion analysis works follow that scheme~\cite{menegola2017recod,nasr2016melanoma,menegola2017knowledge,bi2017automatic,devries2017skin,perez2018data}. Train-only augmentation is still very common~\cite{yu2017automated,esteva2017dermatologist,lopez2017skin,ge2017exploiting,yang2017novel,vasconcelos2017increasing,diaz2017incorporating}. 

In this work, we evaluate two \textbf{augmentation distortion setups} (TensorFlow/Slim's default for each network, and an attempt to customize it for skin lesions). We always apply augmentation to the train samples, but evaluate the impact of applying or not \textbf{augmentation for test samples}.

Works based on global features or bags of visual words often preprocess the images (some recent works using deep learning do it as well) to reduce noise, remove artifacts (e.g., hair), enhance brightness and color, or highlight structures~\cite{sabbaghi2016deep,jia2017skin,matsunaga2017image,wighton2011generalizing,abbas2013melanoma,yoshida2016simple}. The deep-learning ethos usually forgoes that kind of ``hand-made'' preprocessing, relying instead on networks' abilities to learn those invariances --- with the help of data augmentation if needed. 

On the other hand, segmentation as preprocessing is common on deep-learning for skin lesion analysis~\cite{nasr2016melanoma,yang2017novel}, sometimes employing a dedicated network to segment the lesion before forwarding it to the classification network~\cite{yu2017automated,codella2017deep,diaz2017incorporating}. Those works usually report improved accuracies. In this work, we also evaluate the impact of \textbf{using segmentation} to help classification.

If ad-hoc preprocessing (e.g., hair removal) is atypical in deep-learning, \textit{statistical} preprocessing is very common. Many networks fail to converge if the expected value of input data is too far from zero. Learning an average input vector during training and subtracting it from each input is standard, and performing a comparable procedure for standard deviations is usual. The procedure is so routine, that with rare exceptions~\cite{menegola2017recod,kawahara2016deep}, researchers do not even mention it. In this work, we evaluate the impact of two \textbf{image normalization} schemes: TensorFlow/Slim's default for each network, and an attempt to customize it for skin lesions.

Deep network architectures can directly provide the classification decisions, or can provide features for the final classifier --- often Support Vector Machines (SVM). Both the former~\cite{esteva2017dermatologist,lopez2017skin,harangi2017skin}, and the latter~\cite{sabbaghi2016deep,menegola2017knowledge,codella2017deep,ge2017exploiting} procedures are readily found for skin lesion analysis. We~evaluate those \textbf{choices of adding or not an SVM layer}. Also common, are \textbf{ensemble techniques}, which fuse the results~from several classifiers into a final decision~\cite{menegola2017recod, harangi2017skin, matsunaga2017image, bi2017automatic, devries2017skin}. 

As shown above, the literature on automated skin lesion analysis is vast: even the limited scope chosen for this survey comprises more than a dozen works. Making comparisons across those techniques was, until very recently, next to impossible due to poorly documented choices of datasets, splits, implementation details, or even evaluation metrics~\cite{fornaciali2016towards}. The ISIC Challenge~\cite{codella2017skin} has sharply improved that scenario, by providing standards for data and metrics, and by requiring participants to publish working notes. 

A subtler way ISIC Challenge served the community was by clearly separating validation and test datasets, and by keeping test datasets secret until all evaluations were over. When test sets are open, researchers have a strong tendency to optimize hyperparameters on test (\textbf{hyperoptimize on test}, for short).

Hyperoptimizing on test is not the same as overfitting the training: the latter is a \textit{technical} shortcoming in which one overadjusts one's parameters (or hyperparameters) to \textit{train} (or validation) data, while the former is a \textit{methodological} error in which one adjusts one's hyperparameters using \textit{test} data. Hyperoptimizing on test is also different than \textit{directly} train--test contamination --- mixing train and test samples, or using statistics from the test set in the model --- a crude methodological blunder that most practitioners avoid. Hyperoptimizing on test is a \textit{subtle} methodological error in which one \textit{indirectly} uses information from the test set, by, for example, designing, refining and evaluating different models over a test set, picking the best one and reporting its results in \textit{that same} test set~\cite{salzberg1997comparing}. Since the whole process may stretch over weeks or months, researchers fail to realize they are using privileged information to improve the model. As we will discuss, hyperoptimizing on test leads to overoptimistic estimations of performance. 

\textcolor{black}{As discussed in Section~\ref{sec:intro}, existing art lacks a systematic evaluation of the hyperparameters above. The present work intends to fill this gap.}

%
%
\subsection{ISIC Challenge 2017} \label{sec:challenge}

We documented our participation in Parts 1 and 3 of ISIC Challenge 2017 in our working notes~\cite{menegola2017recod}. In this section, we briefly summarize that participation, our findings, and contrast our aims then and now.

Our team has been working on skin lesion analysis since early 2014~\cite{fornaciali2014statistical}, and has been employing deep learning with transfer learning for that task since 2015~\cite{carvalho2015}. On the other hand, we were tackling skin lesion segmentation for the very first time. Our team reached 1st place in melanoma classification (AUC~= 0.874), 3rd place at keratosis classification (AUC~= 0.943), and 3rd place in overall melanoma/keratosis classification (mean AUC~= 0.908). We also reached 5th place in skin lesion segmentation (Jaccard score~= 0.754). The organizers published a comprehensive summary of the entire challenge~\cite{codella2017skin}.

The experiments presented here are inspired by the findings we made during the challenge, but this paper takes different directions, especially regarding methodology. If during the challenge we tempered scientific rigor with the sportive desire to win, here we stress only the former: this work does not focus on maximizing AUCs, but on broader issues of experimental design and validation for the proposed methods.

%
%

\section{Materials and Methods} \label{sec:methods}

\textcolor{black}{In this section we explain in detail the methodology employed in this paper, comprising the factors present in our experimental design, the hardware/software infrastructure used to run the experiments, the statistical design of the experiments, the datasets employed, and the deep-learning models evaluated.}

\textcolor{black}{Our first --- and more practical --- goal is an exhaustive evaluation of the design of deep-learning models for skin lesion analysis. We focused on ten hyperparameters (a choice motivated by their importance in existing literature, as explained in Section~\ref{sec:sota}). Each of those hyperparameters becomes a factor in our experimental design (Table~\ref{tab:factors}).}

\begin{table*}[t]
\centering
\renewcommand{\arraystretch}{1.04}
\caption{Factors in our experimental designs, with corresponding levels }
\begin{scriptsize}
\label{tab:factors}
\begin{tabular}{clp{13cm}}
\toprule
\textbf{Symbol} & \textbf{Factor} & \textbf{Levels} \\ \midrule
a &    Model &    ResNet-101 v2 \textit{versus} Inception v4 \\
b &    Train dataset & Train split of ISIC Challenge 2017 \textit{versus} Full: Level 1 + ISIC Archive + U. of Porto PH$^2$ + U. of Edinburgh Dermofit \\
c &    Input resolution & Pre-augmentation resolution --- 299$\times$299 pixels (305$\times$305 if using segmentation) \textit{versus} 598$\times$598 pixels \\
d &    Train augmentation &    TensorFlow/Slim's default \textit{versus} Level 1 + rotations = on, fast\_mode = off, minimum\_area = 0.20  \\
e &    Input normalization & TensorFlow/Slim's default \textit{versus} Subtract mean of samples' pixels \\
f &    Segmentation &    No segmentation information  \textit{versus}  Segmentation pre-encoded at input \\
g &    Training length & Short (about half the length of Full)  \textit{versus}  Full (30k batches for ResNet / 40k batches for Inception; batch\_size = 32) \\
h &    SVM decision layer & Absent \textit{versus}  Present \\
i &    Test augmentation post-deep & No (decision on single non-augmented sample) \textit{versus} Yes (decision on average of 50 random-augmented samples) \\
j &    Test dataset & Split of Train/Full \textit{vs.} Validation of ISIC Chall. '17 \textit{vs.} Test of ISIC Chall. '17 \textit{vs.} EDRA/Dermoscopic \textit{vs.} EDRA/Clinical  \\
t & Transfer learning & Training from scratch (weights initialized at random) \textit{vs.} Transfer from ImageNet (checkpoint  published by Tensorflow/Slim) \\
\toprule
\end{tabular}
\end{scriptsize}
\end{table*}

Most of those factors are not particular to skin lesion analysis, but are relevant for all image classifiers using deep learning. However, a preoccupation with resolution (c), augmentation customization (d), and segmentation (f) makes more sense for skin lesion analysis --- or at least for medical images in general --- than for general-purpose tasks, like ImageNet.

Our second, more philosophical, goal is discussing methodological issues in the design and evaluation of classification models, especially those which, like skin lesion analysis, aim at medical applications. We are far from the first to point out~\cite{salzberg1997comparing,torralba2011unbiased} that the common practice of meta-optimizing a technique on the same test set used to evaluate the technique leads to over-optimistic results. In this paper, we showcase that effect as we cross-analyze the results in five different test sets.

\subsection{Software and Hardware} \label{sub:materials}

We ran the experiments on a cluster of Ubuntu Linux machines, on a variety of NVIDIA GPUs, including Titan X Maxwell, Titan X Pascal, and Tesla K40. We built the classification models on Python/TensorFlow v.1.3. using the Slim framework. Slim provides ready-to-use models for ResNet-101-v2 and Inception-v4, which we used, with slight adaptations. The statistical analyses ran in R.

In this text we aimed to provide the details needed to understand all results; for complete reproducibility, we provide the end-to-end pipeline, from the raw data to the tables and figures, at our code repository.

\subsection{Experimental Design} \label{sub:design}

The \textbf{main experimental design} was a two-level full factorial design for nine of the ten factors mentioned above (a--i), for each one of the five test datasets (j), resulting in $2^9\times5=2560$ treatments evaluated. We ran a second experiment to evaluate the impact of \textbf{transfer learning}, evaluating seven factors (a--e, g, i, t),  and fixing (f) as no segmentation and (h) as SVM layer absent, resulting in $2^8\times5=1280$ treatments evaluated. In all experiments, we used the \textcolor{black}{AUC} as main metric. Following the ISIC Challenge 2017, we use the mean AUC between the melanoma-vs-all and the keratosis-vs-all as the measured outcome in all experiments.

The analysis for both experiments was a classical multi-way ANOVA, in which the test datasets entered as one of the factors. That choice highlights our aim to make decisions that generalize across datasets, in contrast to maximizing the performance for a particular dataset. 

ANOVA provides both a significance analysis (p-value), and a partition of the variance, which allows to roughly estimate the relative importance/influence of each factor, or combinations of factors. For effect size/explanatory power, we use the $\eta^2$ measure, which is the ratio of the variances (sums-of-squares), extensively used due to its simplicity. The ANOVA table of the main experiment is summarized in Table~\ref{tab:anova} and explored in the next section. We also employed correlograms to highlight issues relating to the choice of the test dataset, and performance metric (\figurename~\ref{fig:correlations}). 

Most of the time, our full factorial designs are too costly to use --- thus our next set of experiments, exploring ensemble techniques, helps in more practical situations. We evaluated a \textbf{straightforward ensemble}, which just pools the decision of several classifiers, and showed that it provides very good performances, without the costs of a full design. 

We also simulated the most common procedure employed by researchers and practitioners: sequential optimization of hyperparameters, in which one starts from a given configuration of hyperparameters, selects one of them to evaluate, commits to the best results, and proceeds to evaluate the next. Although that procedure is very fast (it allows optimizing the nine factors our main design in just 18 experiments), it is sub-optimal in comparison to ensembles. 

Finally, we showed that the customary procedure of \textbf{optimizing the hyperparameters} in the same test set used to evaluate the technique leads to overoptimistic results in both the ensemble and the sequential design. 

Details and results of all procedures appear in Section~\ref{sec:experiments}.

\subsection{Data} \label{sub:data}

Due to deep-learning greediness for data, we sought all high-quality publicly available (for free, or for a fee) sources to compound our dataset: 
      
\begin{enumerate}
    \item \textit{ISIC 2017 Challenge}~{\normalfont\cite{codella2017skin}}, the official challenge dataset, with 2,000 dermoscopic images (374 melanomas, 254 seborrheic keratoses, and 1,372 benign nevi). 
    \item \textit{ISIC Archive}, with over 13,000 dermoscopic images.  
    \item \textit{Dermofit Image Library}~{\normalfont\cite{ballerini2013color}}, with 1,300 clinical images (76 melanomas, 257 seborrheic keratoses). 
    \item \textit{PH2 Dataset}~{\normalfont\cite{mendoncca2013ph}}, with 200 dermoscopic images (40 melanomas).  
    \item \textit{EDRA Interactive Atlas of Dermoscopy}~{\normalfont\cite{argenziano2002dermoscopy}}, with 1,000+ clinical cases (270 melanomas, 49 seborrheic keratoses), each with at least two images (dermoscopic, and close-up clinical). 
\end{enumerate}

We used essentially the same data sources we employed during the ISIC 2017 challenge, except for the no longer available dataset created by the Department of Medical Informatics, RWTH Aachen University (cited in our report~\cite{menegola2017recod} as ``IRMA Dataset''). Even with that exclusion, the new dataset grew, due to better matching diagnostics among the sources (instead of dropping the doubtful cases).

Besides that matching, which we performed with a thesaurus of the terms used in the different datasets, we also annotated images by case, by aliases (same image with different names), and by near-duplicates (two almost-copies of the same lesion). When creating the train and test sets, we barred any cases, aliases, or near-duplicates from splitting across sets.

\begin{figure}[ht!]
    \centering
    \addtolength{\tabcolsep}{-3pt}
    \begin{scriptsize}
    \begin{tabular}{rcccc}
        \renewcommand{\arraystretch}{1.5} %
        \footnotesize
        & Melanoma & Melanoma & Nevus & Keratosis \\
        Dermofit
        & \includegraphics[width=0.085\textwidth, height=0.065\textwidth]{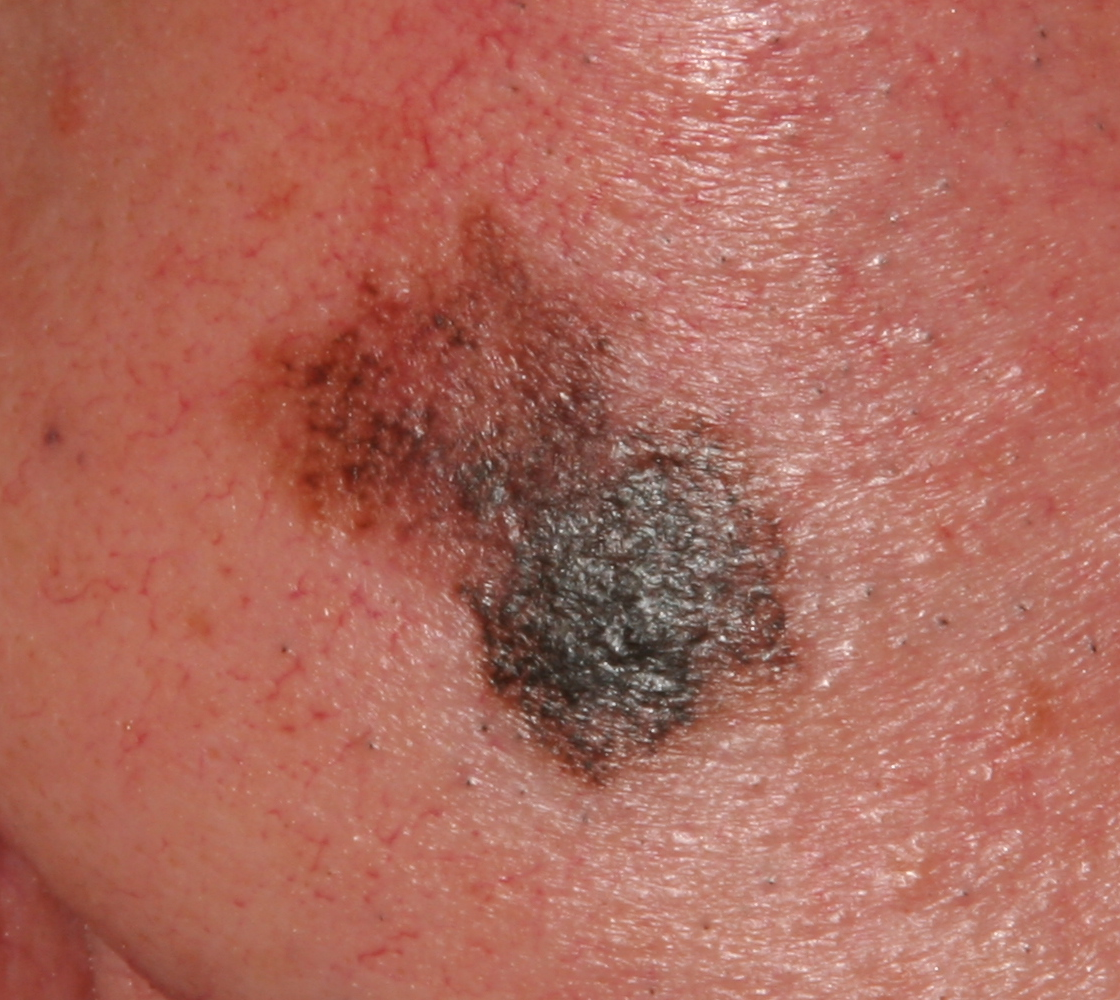} 
        & \includegraphics[width=0.085\textwidth, height=0.065\textwidth]{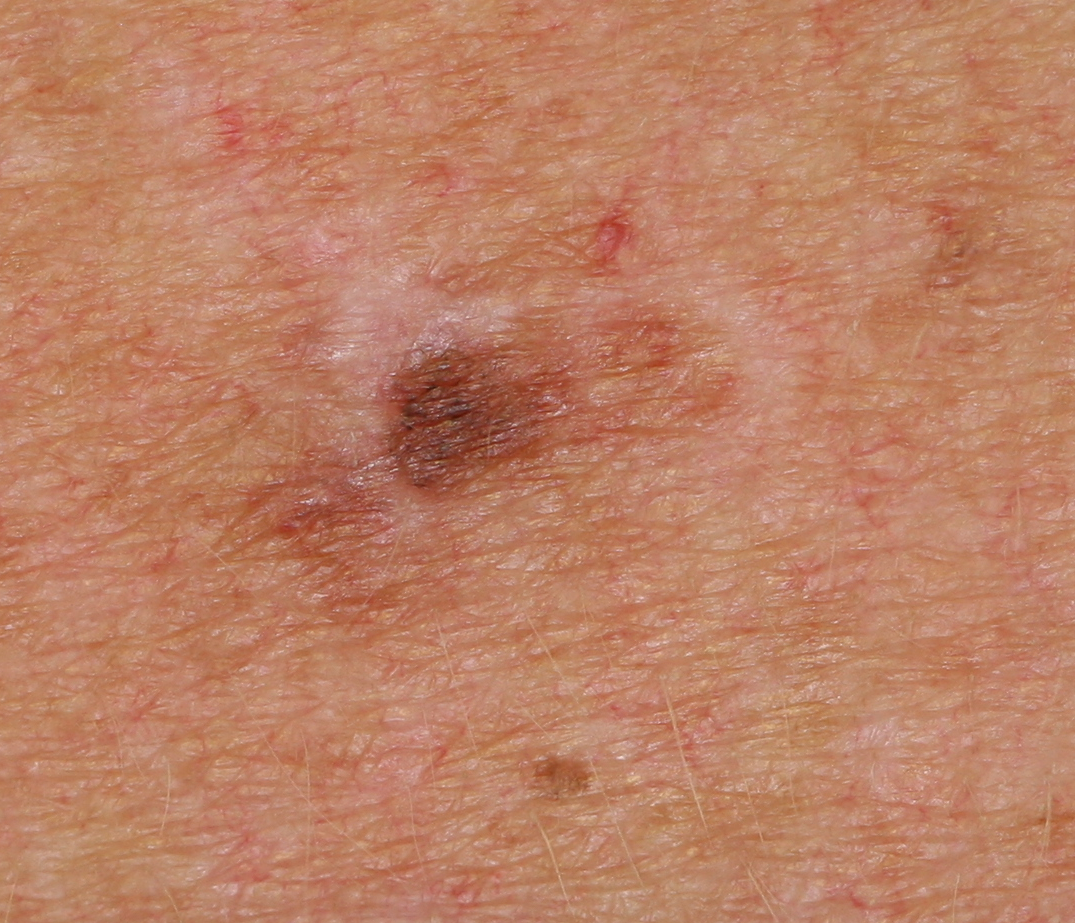} 
        & \includegraphics[width=0.085\textwidth, height=0.065\textwidth]{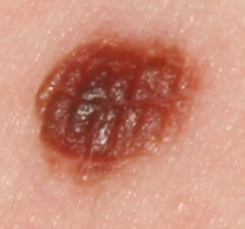} 
        & \includegraphics[width=0.085\textwidth, height=0.065\textwidth]{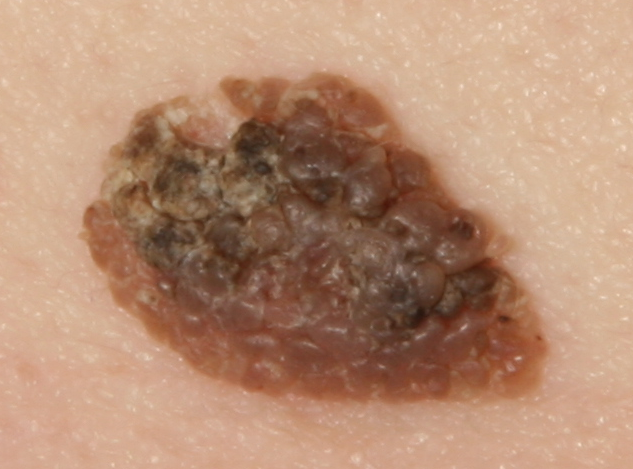} \\
        
        EDRA/Clinic 
        & \includegraphics[width=0.085\textwidth, height=0.06\textwidth]{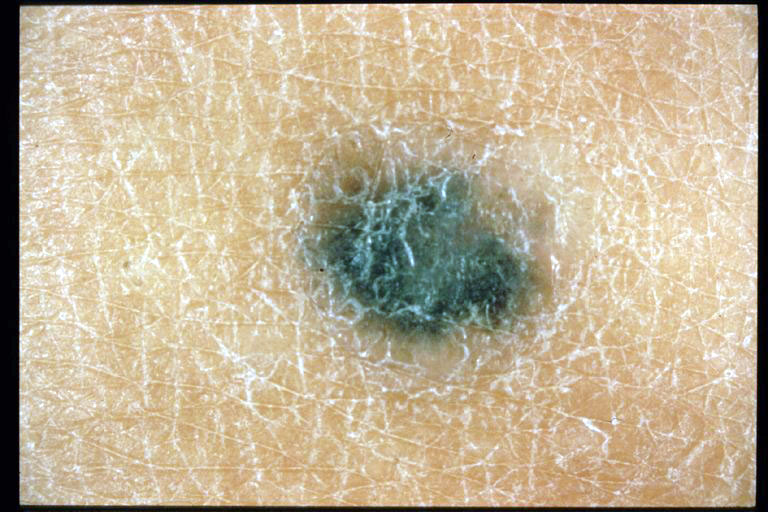}
        & \includegraphics[width=0.085\textwidth, height=0.06\textwidth]{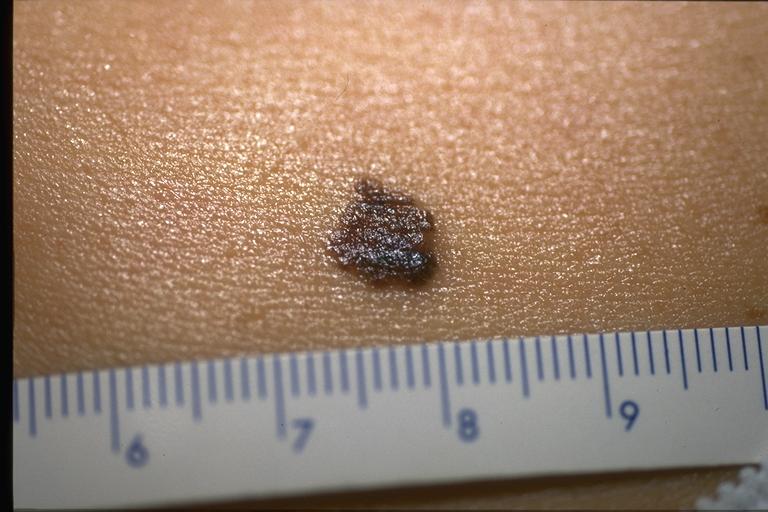}
        & \includegraphics[width=0.085\textwidth, height=0.06\textwidth]{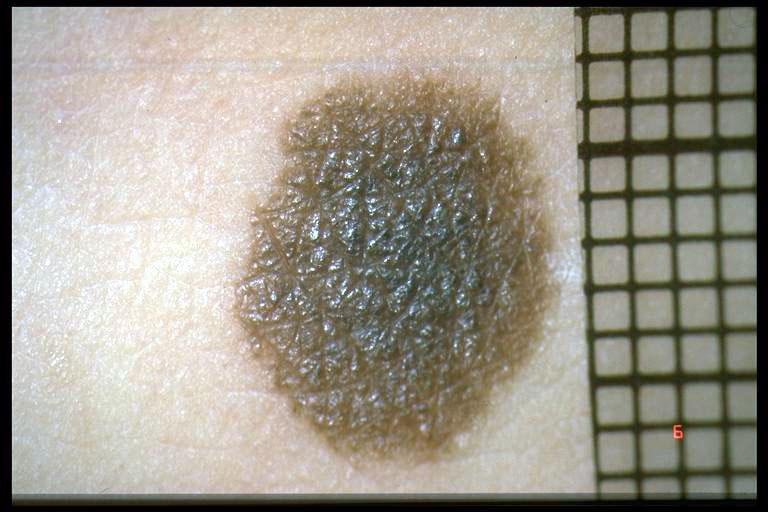}
        & \includegraphics[width=0.085\textwidth, height=0.06\textwidth]{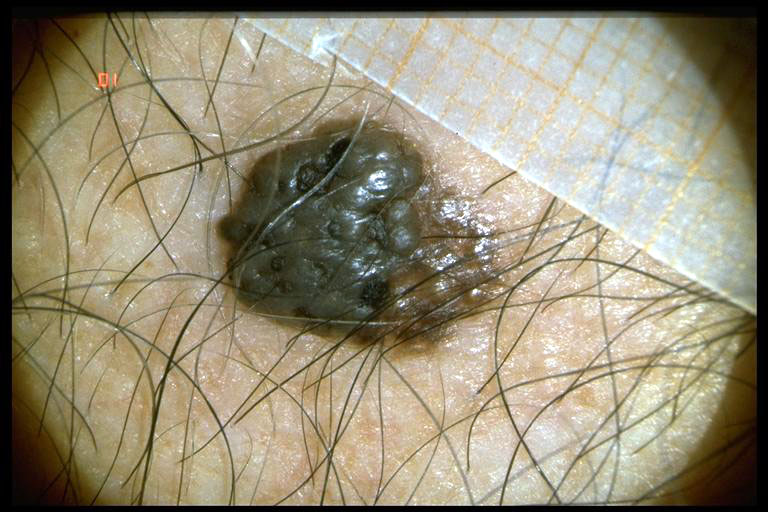}\\
        
        EDRA/Dermato
        & \includegraphics[width=0.085\textwidth, height=0.06\textwidth]{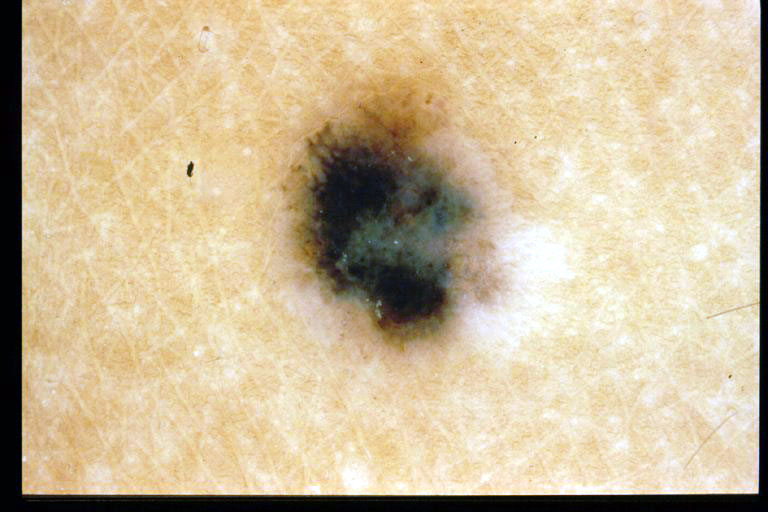}
        & \includegraphics[width=0.085\textwidth, height=0.06\textwidth]{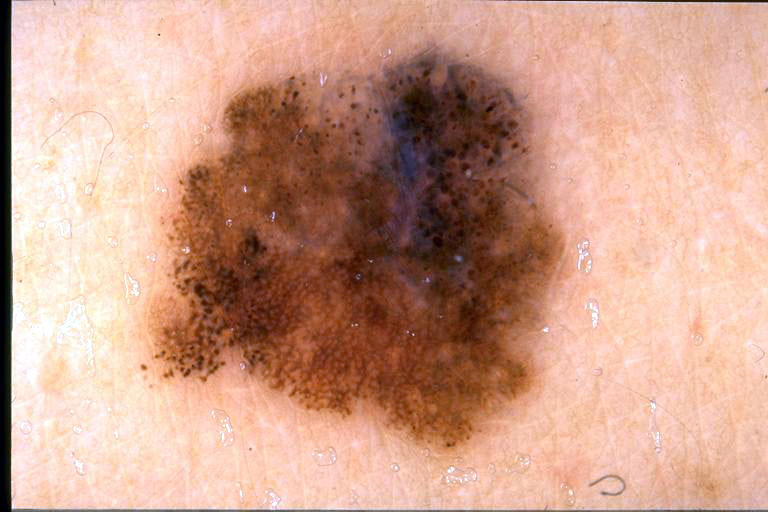}
        & \includegraphics[width=0.085\textwidth, height=0.06\textwidth]{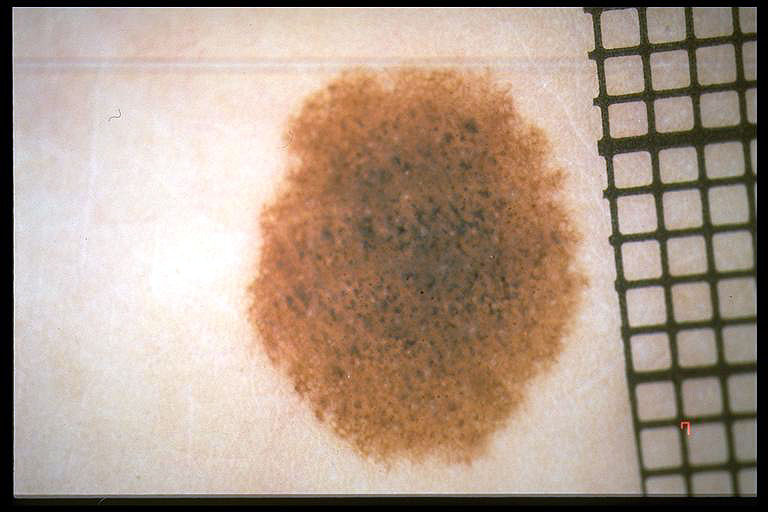}
        & \includegraphics[width=0.085\textwidth, height=0.06\textwidth]{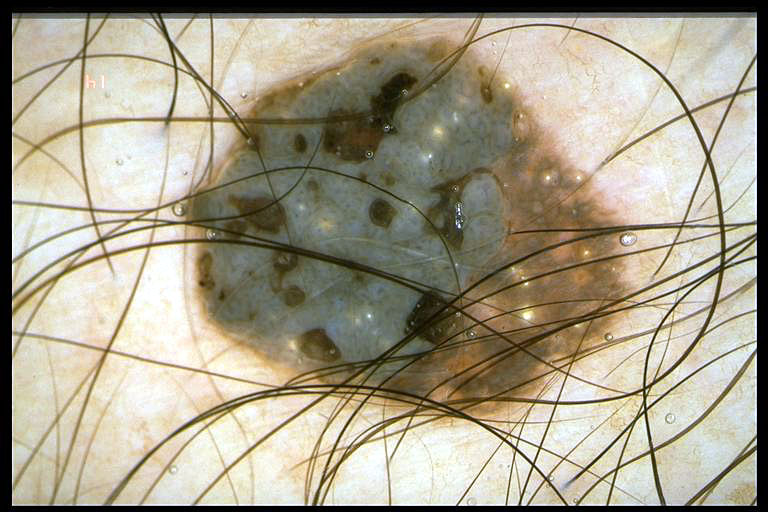}\\
        
        ISIC Archive
        & \includegraphics[width=0.085\textwidth, height=0.06\textwidth]{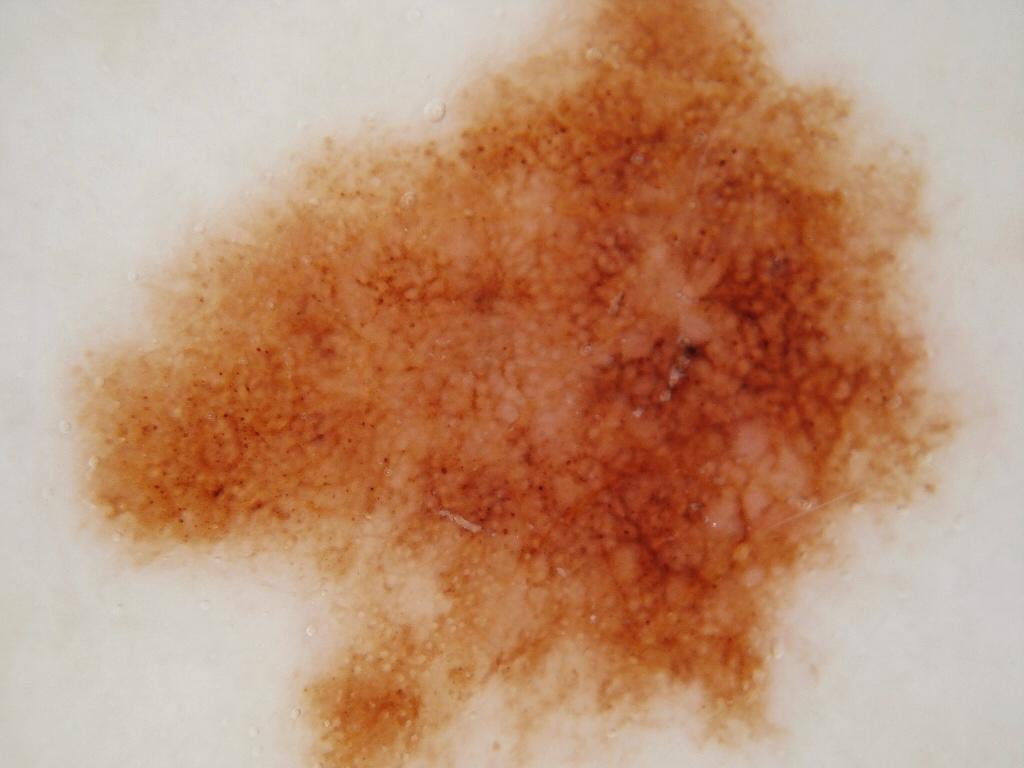}
        & \includegraphics[width=0.085\textwidth, height=0.06\textwidth]{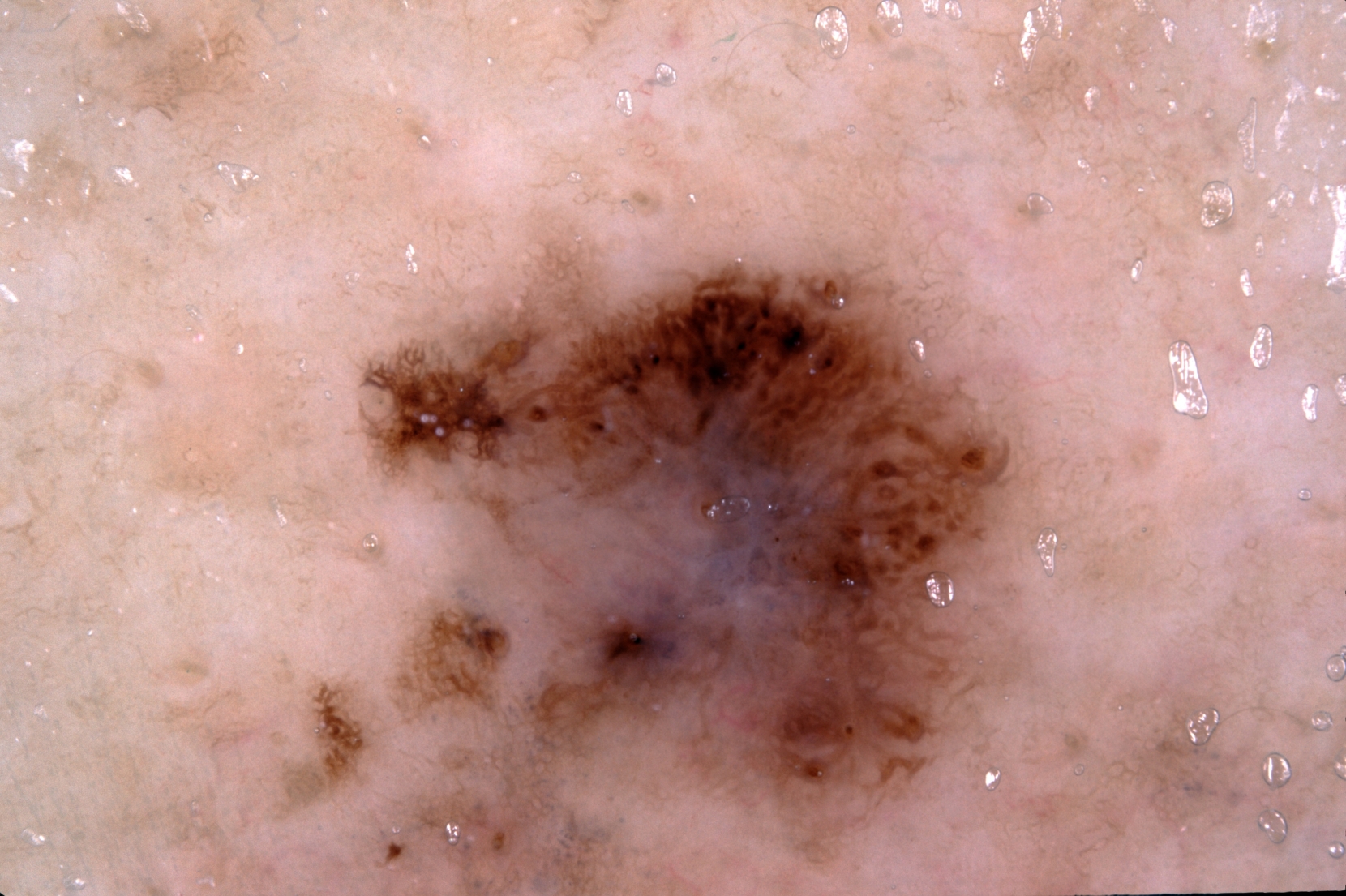}
        & \includegraphics[width=0.085\textwidth, height=0.06\textwidth]{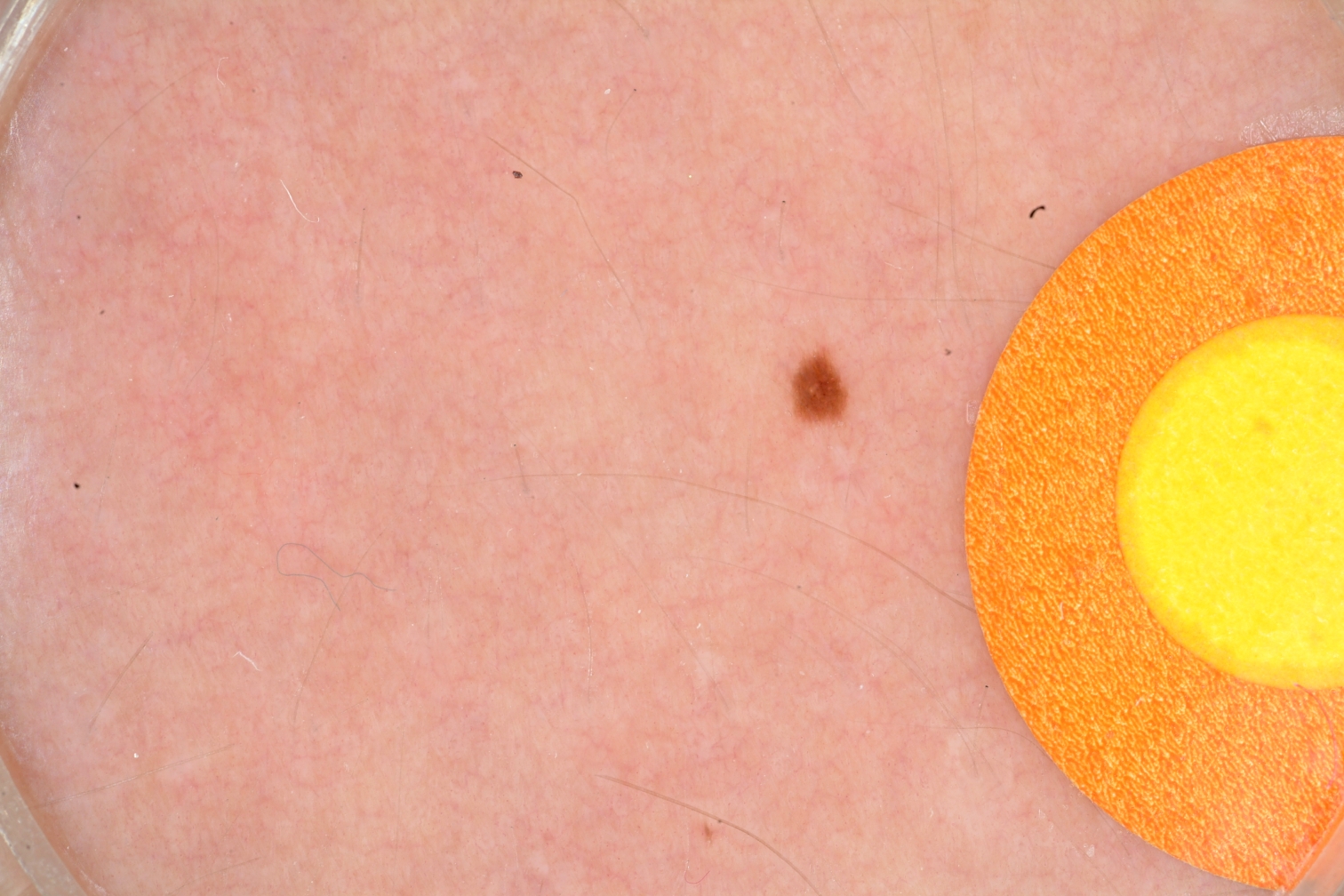}
        & \includegraphics[width=0.085\textwidth, height=0.06\textwidth]{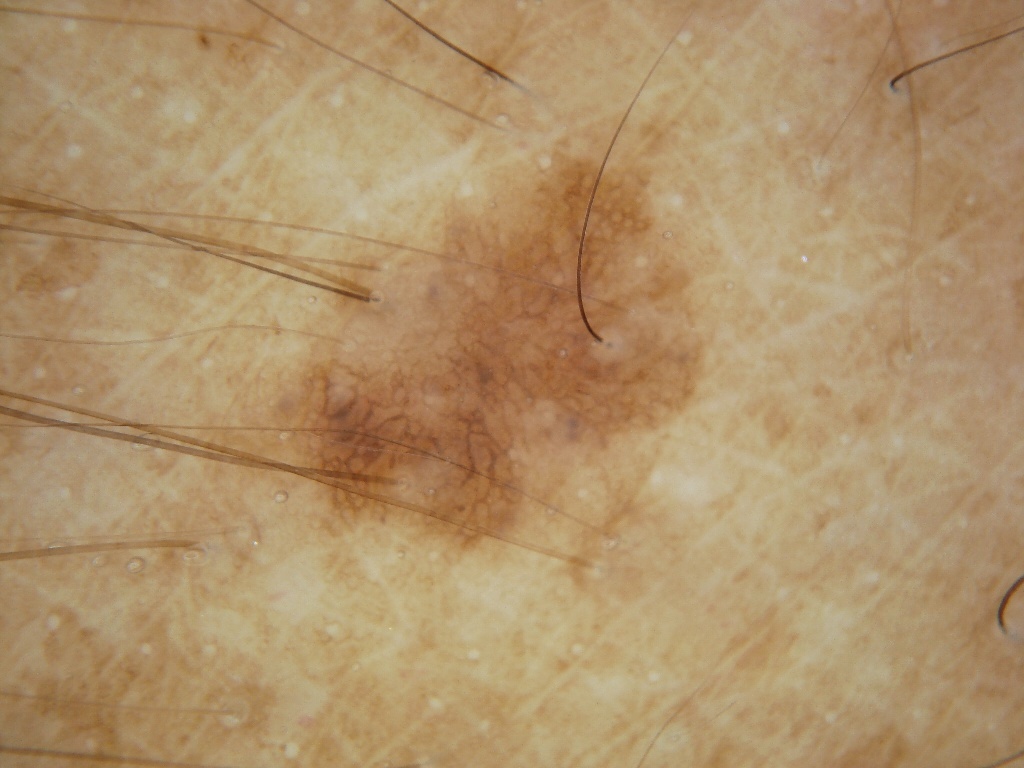}\\
    
        PH2
        & \includegraphics[width=0.085\textwidth, height=0.06\textwidth]{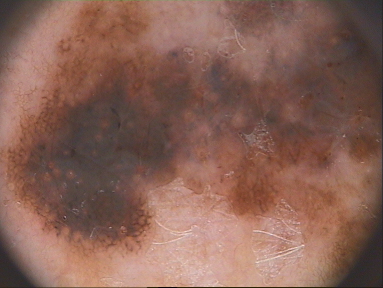}
        & \includegraphics[width=0.085\textwidth, height=0.06\textwidth]{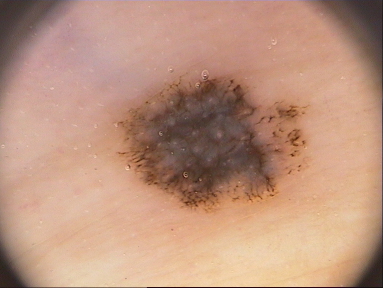}
        & \includegraphics[width=0.085\textwidth, height=0.06\textwidth]{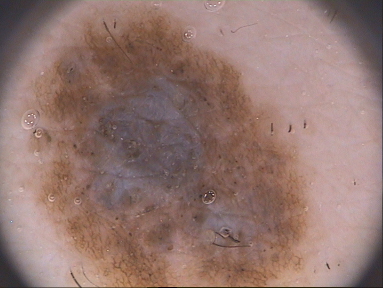} \\
    \end{tabular}
    \addtolength{\tabcolsep}{3pt}
    \end{scriptsize}
    \caption{Samples of skin lesion images. Rows: datasets used in this work. Columns: types of lesions. The first two rows are clinical images; the others are dermoscopic images. Note that the EDRA samples are the same lesion captured in each of the techniques. The PH2 dataset does not have Keratosis lesions. }
    \label{fig:datasets}
\end{figure}

Data sources affect the train and test datasets. For the train dataset (factor b), we contrasted (1) using only the official train split of the ISIC Challenge 2017 dataset, to (2) joining the train split of the ISIC Challenge, the ISIC Archive, the Dermofit Library, and the PH2 Dataset and extracting from that full dataset a train split. 
For the test dataset (factor j), we contrasted (1) an internal test split extracted from our full dataset; (2) the official validation split and (3) the official test split of ISIC Challenge 2017; (4) the dermoscopic images and (5) the clinical images of the EDRA Interactive Atlas of Dermoscopy. Table~\ref{tab:datasets} summarizes the final assembled sets. Figure~\ref{fig:datasets} shows some samples of the datasets. 

\begin{table}[tbp]
\centering
\caption{Summary of the train and test sets.}
\begin{scriptsize}
\label{tab:datasets}
\begin{tabular}{lccc}
\toprule
\textbf{Type}                           & \multicolumn{1}{l}{Melanoma} & \multicolumn{1}{l}{Nevus} & \multicolumn{1}{l}{Keratosis} \\ 
ISIC Challenge 2017 train split         & 374                                    & 1372                                 & 254                                     \\  
Full train (composition of datasets)    & 1227                                   & 10124                                 & 710                                     \\ \cmidrule(lr){1-4}
Internal test split from full           & 135                                    & 3129                                 & 89                                     \\
ISIC Challenge 2017 validation split    & 30                                     & 78                                 & 42                                     \\
ISIC Challenge 2017 testing split       & 117                                    & 393                                 & 90                                     \\ 
EDRA Atlas of Dermoscopy (each version) & 518                                    & 1154                                 & 95                                     \\ 
\bottomrule
\end{tabular}
\end{scriptsize}
\end{table}

\subsection{Models} \label{sub:models}

For the main experiment, we always employed pre-trained models that proved successful for the ImageNet challenge, loading the weights from checkpoints published in TensorFlow repository. For the transfer experiment, we contrasted those with the same models initialized with random weights.

To evaluate the choice of the model (factor a) we contrasted two architectures: ResNet-101-v2~\cite{He2016eccv}, and Inception-v4~\cite{szegedy2016inceptionv4}, using the reference implementation available in TensorFlow/Slim v1.3. 

In this paper, segmentation was used only as an ancillary input for classification (factor f).
For the ISIC Challenge 2017, we had used a segmentation network based on the work of Ronneberger et al.~\cite{RonnebergerFB15} and Codella et al.~\cite{codella2015deep}. For this work, we streamlined that model, reducing the number of parameters, removing the fully-connected and Gaussian-noise layers, and adding batch-normalization and dropout layers. The new model\footnote{https://github.com/learningtitans/isbi2017-part1} is faster to train and occupies much less disk space. We trained the segmentation models on the same images as their corresponding classification models.

Because of the lack of literature consensus on how to use segmentation for skin lesion analysis, we opted for schemes with minimal changes to both data and networks. We pre-evaluated two candidates: pixel-wise multiplying the input RGB images by the segmentation masks \textit{versus} pre-encoding the four planes (\textcolor{black}{Red, Green, Blue} and mask) into three planes, keeping the rest of the networks unchanged. For the full design, we only considered the latter, which appeared more promising on those preliminary tests.

Pre-encoding the masks required slightly adapting ResNet and Inception, by adding the pre-encoding adapter layers. For both ResNet/Inception we added three convolutional layers before the input, two layers with 32 filters, and a third with 3 filters. All convolutional layers used $3\times3$ kernels and stride of 1. Since ResNet-101-v2 and Inception-v4 models require input images of $299\times299$ pixels, the adapter layer took $305\times305$-pixel images, to account for the 2 border pixels lost at each convolutional layer.

%
%
\section{Results and Discussion} \label{sec:experiments}

\textcolor{black}{In this Section we show and discuss the results of the experiments introduced in Section~\ref{sub:design}. For easier understanding, we organized the extensive results into five subsections: hyperparameterization of deep learning for melanoma screening, importance of transfer learning, pitfalls of sequential designs, power of ensembles, and pitfalls of hyperoptimizing on test.}

\textcolor{black}{\subsection{Deep-learning hyperparameterization}} 

As explained, the main experiment was a full factorial design with nine two-level factors (a--i), and five test datasets (factor j). We used a classical multi-way ANOVA with the mean AUC for melanoma and keratosis as the measured outcome (with the small technicality of taking the logit of that measure, since, when working with rates, the logit helps to fulfill ANOVA's assumption of Gaussian residuals). We considered all main effects, and up to 3-way interactions. We considered higher-order interactions unlikely and assigned them to the residuals. 

Table~\ref{tab:anova} shows a summary of main experiment's ANOVA, with the symbols for the factors and interactions on the first column, and the names of the main factors on the second. The remaining columns show the outcomes of the test. The most important columns are \textit{p-value}, which measures statistical significance, and \textit{explanation (\%)}, which measures effect-size/explanatory power. The p-value is inferred, as usual, from the F-statistic of ANOVA, while the explanatory power uses the $\eta^2$ measure. 

\begin{table*}[t]
\centering
\renewcommand{\arraystretch}{1.04}
\caption{Selected lines from the 176-line ANOVA table; most of the omitted lines (126) had p-values $\geq$ 0.05. Absolute explanation based on $\eta^2$-measure, relative explanation ignores residuals and choice of test dataset (j).}
\label{tab:anova}
\begin{tabular}{clccccccc}
\toprule
  & &  &  \multicolumn{2}{c}{Explanation (\%)} & \multicolumn{2}{c}{Best AUC (\%)} & \multicolumn{2}{c}{Worst AUC (\%)} \\ 
      & Factor               & p-value  & Absolute & Relative & Treatment               & Mean & Treatment                   & Mean \\  \toprule
a     & Model architecture   & $<$0.001 & 0        & 1        & resnet                  & 84   & inception                   & 83   \\ 
b     & Train dataset        & $<$0.001 & 5        & 46       & full                    & 85   & challenge                   & 81   \\ 
c     & Input resolution     & $<$0.001 & 1        & 5        & 598                     & 84   & 299--305                    & 82   \\ 
d     & Data augmentation    & 0.17     & 0        & 0        & default                 & 83   & custom                      & 83   \\ 
e     & Input normalization  & 0.001    & 0        & 0        & default                 & 83   & erase mean                  & 83   \\ 
f     & Use of segmentation  & $<$0.001 & 0        & 2        & no                      & 84   & yes                         & 83   \\ 
g     & Duration of training & 0.003    & 0        & 0        & full                    & 83   & half                        & 83   \\ 
h     & SVM layer            & $<$0.001 & 0        & 4        & no                      & 84   & yes                         & 83   \\ 
i     & Augmentation on test & $<$0.001 & 1        & 12       & yes                     & 84   & no                          & 82   \\ 
j     & Test dataset         & $<$0.001 & 75       &          & full.split              & 96   & edra.clinical               & 66   \\ 
a:b   &                      & $<$0.001 & 1        & 8        & inception/full          & 86   & inception/challenge         & 80   \\ 
a:f   &                      & $<$0.001 & 0        & 2        & resnet/no               & 84   & inception/yes               & 82   \\ 
b:e   &                      & $<$0.001 & 0        & 2        & full/default            & 86   & challenge/default           & 80   \\ 
b:j      &                      & $<$0.001 & 2        &          & full/full.split         & 98   & chall/edra.clinical         & 63   \\ 
h:j   &                      & $<$0.001 & 0        &          & no/full.split           & 97   & yes/edra.clinical           & 65   \\ 
i:j   &                      & $<$0.001 & 0        &          & yes/full.split          & 97   & no/edra.clinical            & 65   \\ 
a:b:d &                      & $<$0.001 & 0        & 2        & inception/full/custom   & 86   & inception/challenge/custom  & 78   \\ 
a:d:e &                      & $<$0.001 & 0        & 2        & resnet/custom/default   & 85   & inception/custom/default    & 81   \\ 
a:f:j &                      & $<$0.001 & 0        &          & resnet/yes/full.split   & 97   & inception/yes/edra.clinical & 65   \\ 
b:d:e &                      & $<$0.001 & 0        & 1        & full/custom/default     & 86   & challenge/custom/default    & 79   \\ 
c:e:f &                      & $<$0.001 & 0        & 1        & 598/default/no          & 86   & 299--305/default/yes             & 82   \\
& Residuals    & --- & 12    \\ \toprule                        
\end{tabular}
\end{table*}

We present the absolute explanation (considering the entire table) for reference, but our analysis is focused on the relative explanation, which ignores the choice of the test set (j) and the residuals. The reason for ignoring those is that they are not actual choices for designing a new model; therefore, relative explanations indicate better the relative importance of choices to practitioners.

The original full table contained all main effects, and up to 3-way interactions. However, not surprisingly, 126 of the resulting 176 lines were non-significant interactions, which were omitted here. We also left out those interactions with relative explanations lower than 1\%, even if significant. With the notable exception of the customized data augmentation (d), all main effects were significant, but most of their relative explanations were small. 

The analysis of the relative explanation shows an unsurprising, but still disappointing result: the performance gains (b) are almost wholly due to the usage of more data. Other than data, the most important factor was the use of data augmentation on test (d). We performed it, as usual, by taking the test image, generating a number (in our case, 50) of augmented samples exactly like in training, collecting the prediction for each of the samples, and pooling the decisions (in our case, by taking the average prediction). Although not surprising for the literature of deep learning, that finding is relevant for the literature of skin lesion analysis, where many works still forgo augmentation in the test.

Most of the findings tended to confirm the (limited) observations we made during the ISIC Challenge 2017, with two notable exceptions. Input resolution (c), which we deemed unimportant during the challenge, turned out to have a non-negligible effect. That result is particularly interesting, because we used a very rough form of augmented resolution, by inputting high-resolution images to the augmentation engine, but still feeding normal-resolution crops to the network. On the other hand, the use of an SVM decision layer (h), which we considered advantageous during the Challenge turned out to have a large-effect... only negative! Globally, ANOVA shows it is better \textit{not} to use the SVM.

Normalization (e) and training duration (g) showed tiny ($<$1\%), but still significant positive effects. The choice for those factors must consider their very different costs: adding normalization costs next to nothing, both in implementation complexity and in training time. Training duration doubled the already many-hours-long training times.

As usual, most of the interactions were not significant, and even the ones that were, had effect sizes too small to be worth noting. A notable exception was the interaction between model architecture and train dataset (a:b), whose 8\% of relative explanation was bigger than most main effects. Model choice alone favors the simplest ResNet over Inception, but the combination of Inception with the full dataset is so advantageous that it offsets that effect. We had already observed, informally, this synergy between more data and deeper models during the Challenge.

The most disappointing result was the use of segmentation, which was more than unhelpful, harmful. This result, however, is contingent on our choice for adding segmentation to classification.

\textcolor{black}{\subsection{The importance of transfer learning}} 

We ran a second full factorial design, with seven of the ten factors of the main experiment (a--e, g, i, j), fixing factors (f) and (h), and adding a factor to evaluate the presence \textit{versus} absence of transfer learning (factor t). The new factorial design, with $2^8\times5=1280$ treatments, shows  transfer learning as critical for performance: it explains (favorably) 14.7\% of the absolute variation, and a whopping 62.8\% of the relative variation of performance (computing those metrics the same way as the in the main experiment, i.e., excluding the residuals, and the choice of test dataset and its interactions from the relative variation), with high significance (p-value below 0.001). We omit the ANOVA table for concision. Those results reinforce previous findings on the importance of transfer learning for skin lesion analysis~\cite{menegola2017knowledge}.

\textcolor{black}{\subsection{The pitfalls of sequential designs}} 

As mentioned, full factorial designs are way too expensive for the majority of situations. The most common procedure is the exact opposite: taking a single factor to optimize, and performing a couple of experiments on that factor alone, keeping all others fixed (starting from a combination considered reasonable). Once a factor is decided, one commits to it and takes the next to optimize, until the procedure is complete.

We evaluate the impact of such sequential procedure, simulating it using the measurements on the full design. We take, at random, both the starting treatment and the sequence of factors to test. For factors not yet optimized, the level is given by the starting treatment. Each factor is optimized in turn, by comparing the performance of the alternative treatments on the full-factorial data of a chosen hyperoptimization dataset. The outcome of a single simulation is the performance of the optimized treatment on a chosen measurement dataset. We use the mean keratosis/melanoma AUC as the performance metric. 

\begin{figure*}[t]
    \centering
    \subfloat[Simulated sequential design on ISIC test split.]{\includegraphics[width=0.45\textwidth]{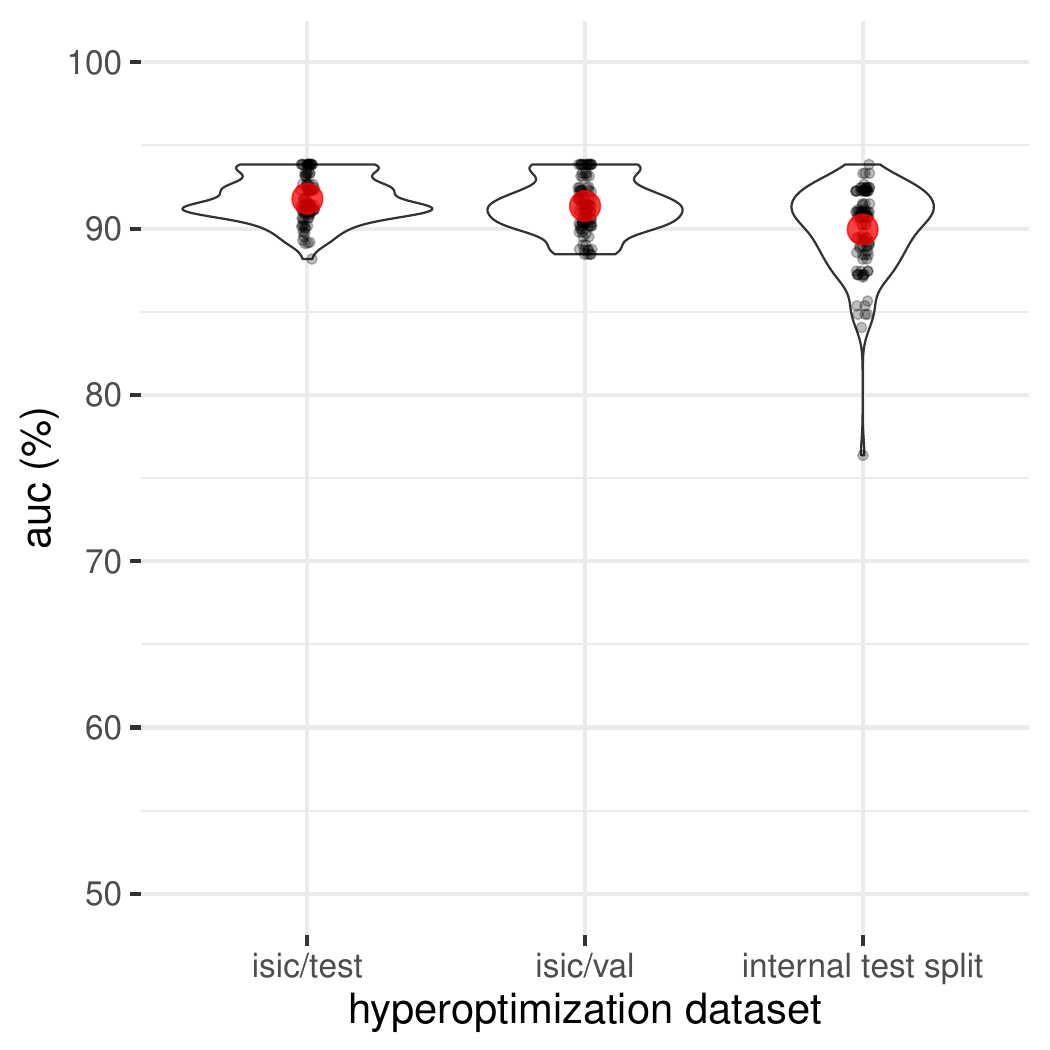}\label{fig:sequential-a}} \hspace{0.2cm}
    \subfloat[Simulated sequential design on EDRA Atlas clinical images.]{\includegraphics[width=0.45\textwidth]{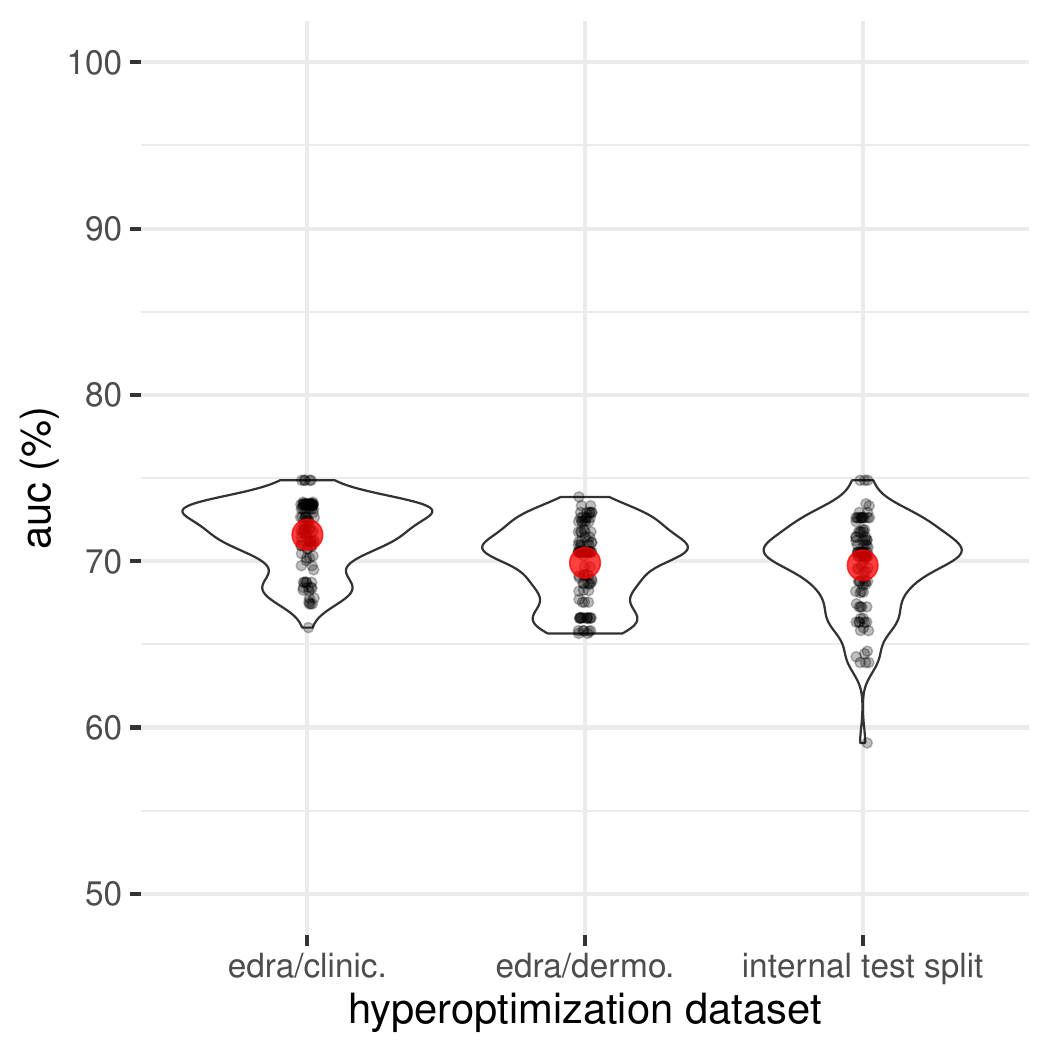}\label{fig:sequential-b}} 
    \caption{Simulation of the sequential optimization of hyperparameters, considering all nine factors (a--i) as the main full factorial. Factors optimized on the dataset shown on the horizontal axis, and performance (mean melanoma/keratosis AUC) measured on the dataset indicated on the captions. For each case, we run 100 simulations, with random optimization sequences and starting points. \changed{The violin plots show the kernel density estimation of the actual data (black dots) and the large red dot shows their mean.}}
    \label{fig:sequential}
\end{figure*}

\figurename~\ref{fig:sequential} shows the results for pairs of hyperoptimization $\times$ measurement datasets, where we perform 100 simulations for each pair. The actual measurements appear as black dots, and the violin plots show their estimated density, while the big red dot shows their mean. \textcolor{black}{The most important observation here is the large variability of the results, visible in both the dispersion of the actual observations (black dots) and the spread of the estimated distribution (violin plots). That variability showcases the instability of the sequential procedure, which depends both on the order in which one optimizes the factors, and on the starting value for them.}

Another notable observation is the (unrealistic) advantage of hyperoptimizing and measuring on the same dataset: not only do we get higher averages, but also a smaller variability. The advantage of hyperoptimizing and measuring on splits of the same dataset is more subtle, but present. \textcolor{black}{That phenomenon will be discussed in more detail in Section~\ref{sec:hyperoptimizing_on_test}.}

\begin{figure*}[t]
    \centering
    \subfloat[Correlogram of mean melanoma/keratosis AUC across test datasets.]{\includegraphics[width=0.45 \textwidth]{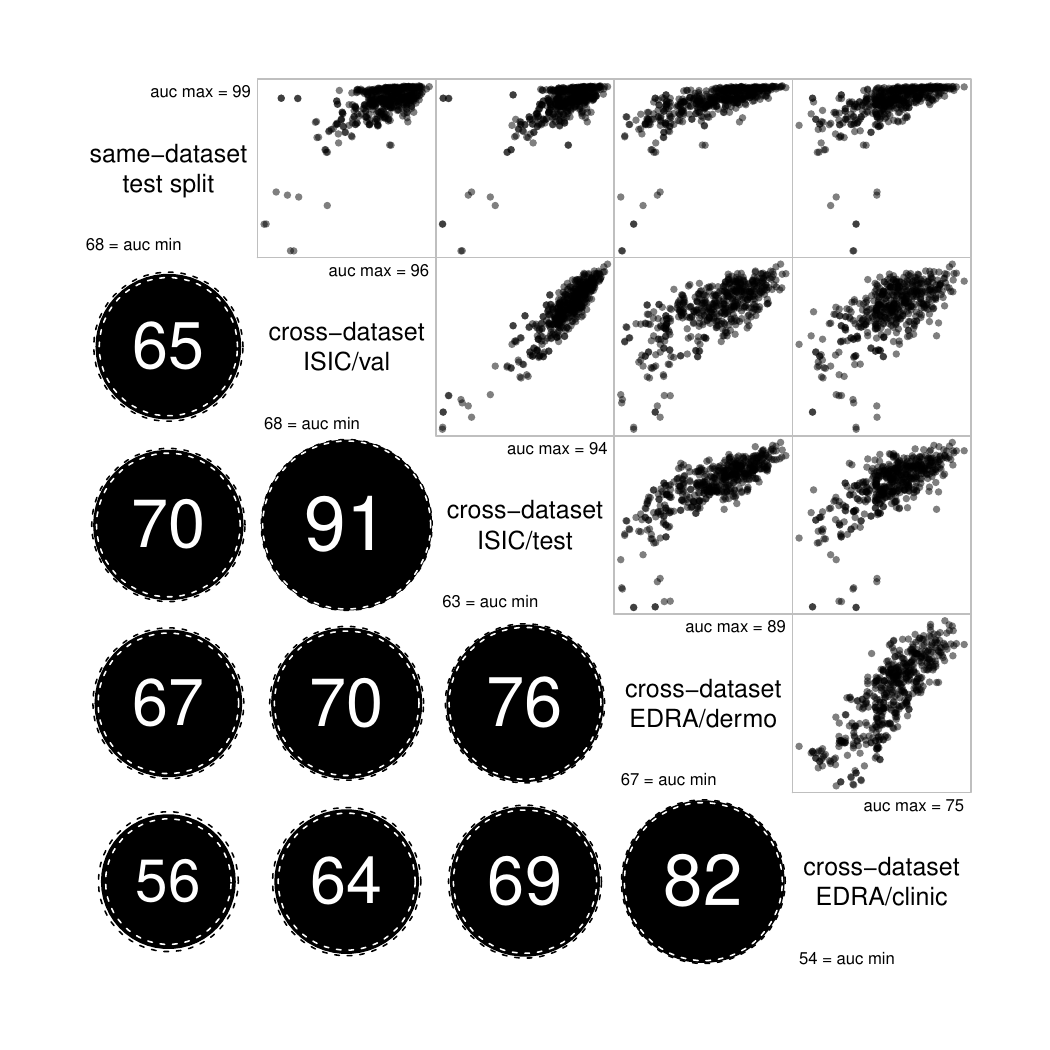}\label{fig:correlations-a}} \hspace{0.2cm}
    \subfloat[Correlogram of metrics on ISIC 2017 Test dataset across metrics.]{\includegraphics[width=0.45 \textwidth]{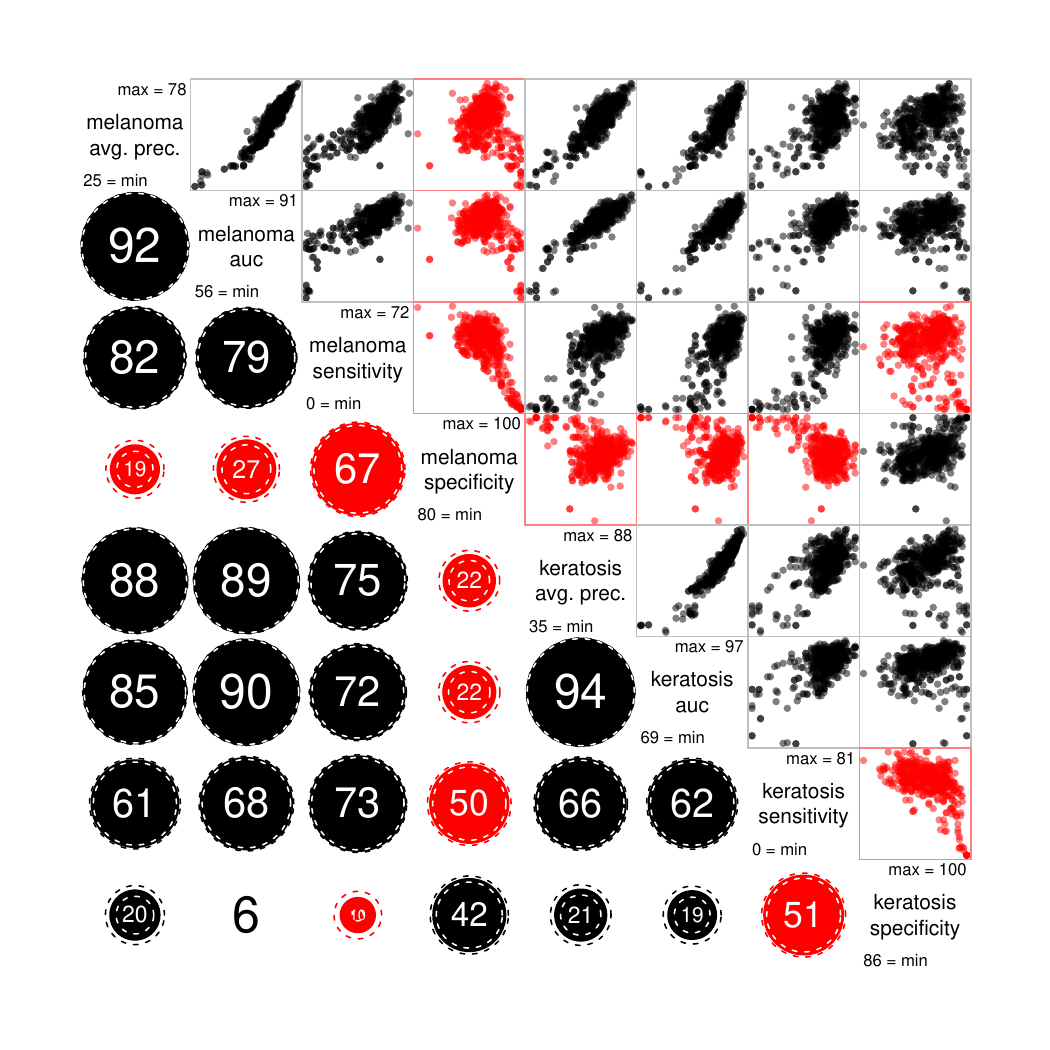}\label{fig:correlations-b}} 
    \caption{Correlograms with pair-wise correlation analyses. Sets appear on the diagonal; upper matrices show the scatter plots, and lower matrices show the Spearman correlation of each pair of sets. On lower matrices, numbers and solid circles' areas show the mean estimates, and dashed circles' areas show the 95\%-confidence bounds. Non-significant estimates appear without the circles. All numbers in \%, negative correlations are in red.}
    \label{fig:correlations}
\end{figure*}

\textcolor{black}{\subsection{The power of ensembles}} 

The expense of the full factorial design, the instability of the sequential procedure, and the limited correlation of performances across datasets seem to leave few options to practitioners. Fortunately, single-model schemes are seldom used today, and ensembles of several models help to alleviate those issues.

We simulated different ensemble strategies, by pooling the predictions of models present in our full design. We evaluate three pooling strategies: average, max, and extremal. Average- and max-pooling work as usual. Extremal pooling takes, from the list of values being pooled, the value most distant from 0.5 --- it may be seen as an ``hypothesis-invariant'' max-pooling. In all cases, after pooling, we re-normalize the probability vector to ensure it sums up to one. Half the models in the full design entered as candidates, and we discarded in this experiment the models with the SVM layer, due to issues in making their probabilities commensurable with the deep-only models.

\begin{figure*}[t]
    \centering
    \includegraphics[width=\textwidth]{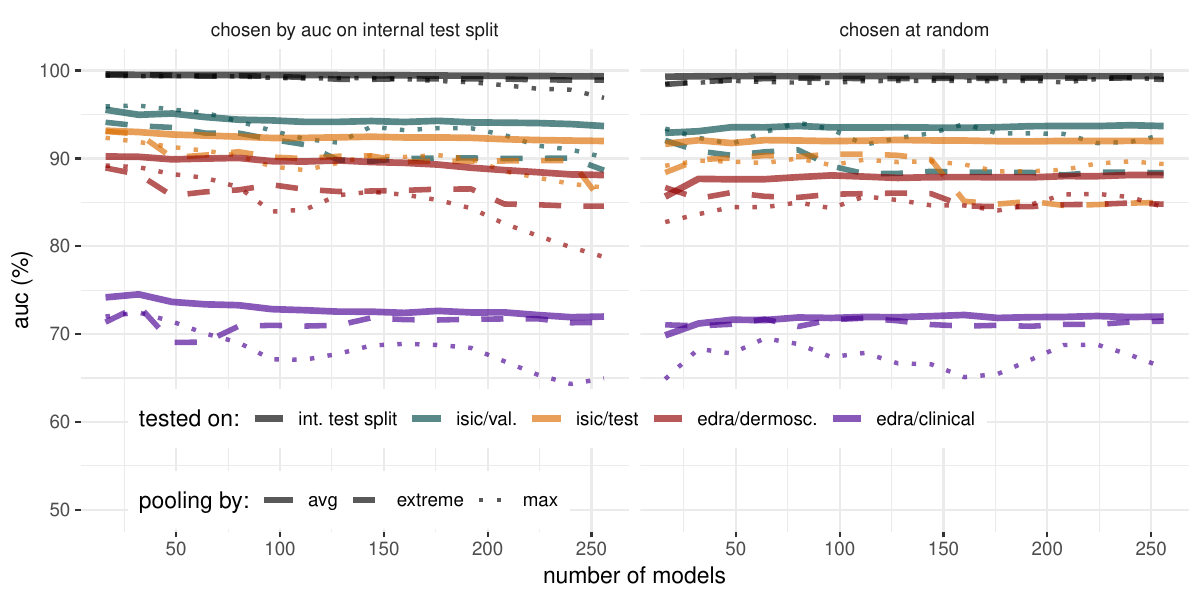}
    \caption{Evaluation of ensemble strategies, by pooling the prediction of a given number of partial models using average-, max-, or extremal-pooling. Left: cumulative effect of adding partial models, starting with the best (as evaluated by the internal test split). Right: same plot, with models randomly shuffled. (Best viewed in color.)}
    \label{fig:ensemble}
\end{figure*}

\figurename~\ref{fig:ensemble} shows the main results. Average-pooling was, by far, the best choice for pooling the decision. Such clear-cut advantage came as a surprise for us, as max-pooling often outperforms average-pooling in related tasks. If no other information is available, simply average-pooling randomly selected models is a reasonable strategy. 

\begin{figure*}[ht!]
    \centering
    \subfloat[Ensembles tested on ISIC 2017/testing.]{\includegraphics[width=0.49\textwidth]{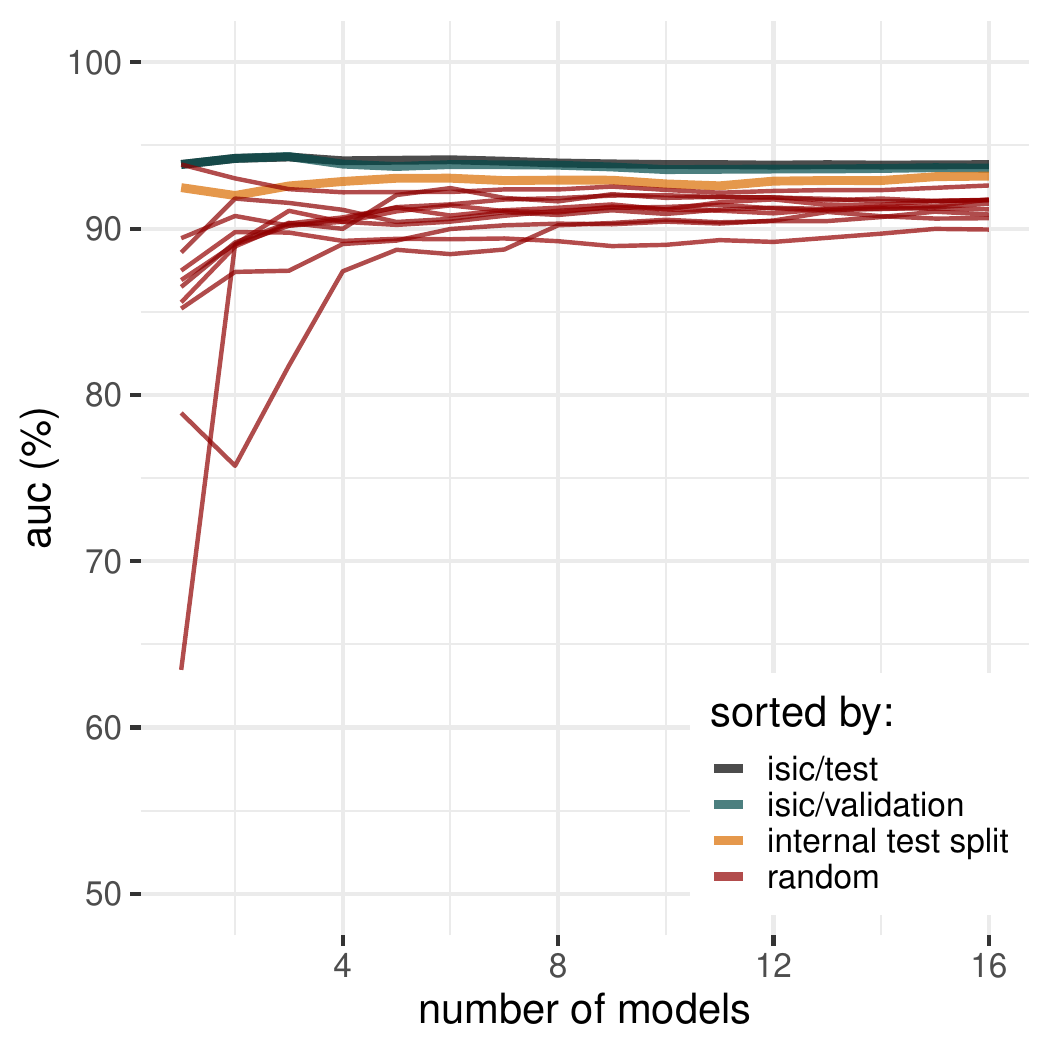}\label{fig:zoomed_meta-a}}\hspace{0.1cm}
    \subfloat[Ensembles tested on EDRA Atlas/clinical.]{\includegraphics[width=0.49\textwidth]{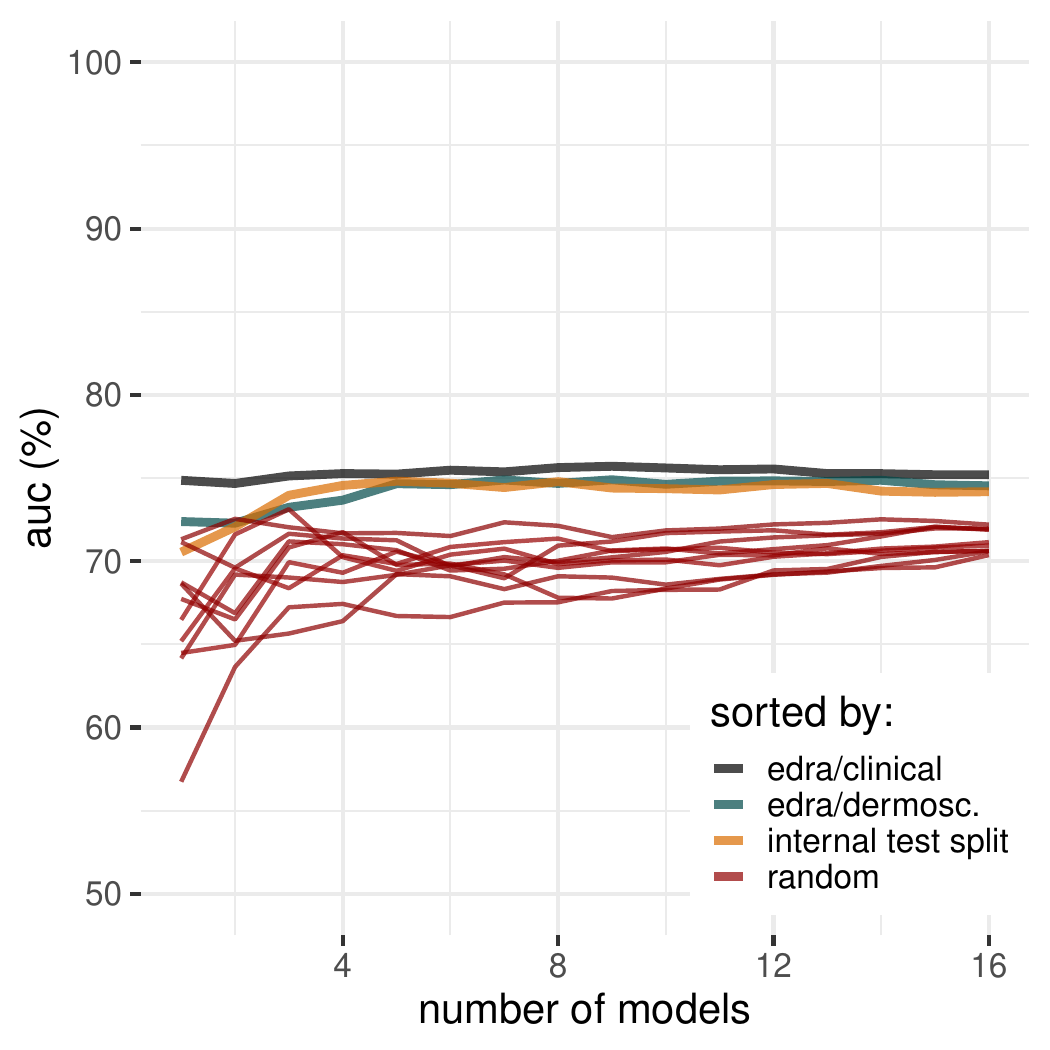}\label{fig:zoomed_meta-b}} 
    \caption{Detailed analysis of ensembles, contrasting the dataset used to choose the models (shown as different curves) in order to optimize the results measured on the test dataset indicated in the captions. We sampled 10 different random ensembles. (Best viewed in color.)}
    \label{fig:zoomed_meta}
\end{figure*}

The use of dozens --- even hundreds --- of models may sometimes be justified in critical tasks (like medical decisions), but training and evaluating so many deep networks is cumbersome. Fortunately, as \figurename~\ref{fig:zoomed_meta} shows, a handful of models seem to work just as well. The results shown here are the ``good news'' part of this paper: we can escape the expense of the full factorial design, and the instability of the sequential designs, by averaging a dozen or so models with parameters chosen entirely at random --- although the random ensembles start very unstable, they soon converge to a reasonable model, in average and variability. If we decide to perform a full factorial, there is good news too: the best models learned in one dataset seem to be informative to compose the ensembles in other datasets, allowing to get top performances with very small ensembles. 

Here, again, the unfair advantage of optimizing (selecting the models for the ensemble) and measuring performance on the same dataset appears. The advantage is small but systematic for the test split of ISIC (\figurename~\ref{fig:zoomed_meta-a}); it is much more apparent for the challenging collection of clinical images of EDRA Atlas (\figurename~\ref{fig:zoomed_meta-b}).

\textcolor{black}{\subsection{The pitfalls of hyperoptimizing on test}} \label{sec:hyperoptimizing_on_test}

As a final experiment, we made two simulations of ``new'' submissions to the ISIC 2017 Challenge. The first simulates a \textit{blind} procedure, mimicking the conditions of the challenge (limited information about the validation split, no information whatsoever about the test split). In that simulation, we assume a full factorial experiment performed only on our internal validation split, and keep the 32 best models (as measured in that split). We then test 32 incremental ensembles on the ISIC validation set, finding that the ensembles with 15 or 16 models have the best performance. We commit to the ensemble with 15 models and do \textit{one} evaluation of that ensemble on the ISIC test split, finding AUCs of  0.895 (melanoma), 0.967 (keratosis), and 0.931 (combined). To put those numbers in perspective, in the actual competition the best AUCs were, respectively, 0.874, 0.965, and 0.911 (obtained by different participants). 

We also simulated a \textit{privileged} procedure, hyperoptimizing without restraint on the ISIC Challenge test split itself. We use the full design on the ISIC Test split to select the 32 best models, and then test 32 incremental ensembles, handpicking the best result for melanoma and for keratosis. We find AUCs of 0.916 (melanoma), 0.970 (keratosis), and 0.943 (combined). Again, to get a better grasp of the difference, consider that the 2.p.p. increase on the melanoma AUCs is larger than the difference between the 1st and the 4th teams ranked in the 2017 Challenge.

\textcolor{black}{The results above are reinforced by} additional correlation analysis with the full factorial experiment (\figurename~\ref{fig:correlations}), which highlight the correlations (a) among results on different test datasets; and (b) among different metrics. To keep the scatter plots directly interpretable, instead of taking the logit of the rates, we dealt with the non-linearity by using Spearman's $\rho$ instead of Pearson's $r$ as correlation measure.

The correlogram on \figurename~\ref{fig:correlations-a} employs the same metric as the ANOVA, the mean melanoma/keratosis AUC used as main metric at the ISIC 2017 Challenge. The test dataset names appear in the diagonal, along with the maximum and minimum AUCs obtained for the 512 variations of the full design on that dataset. The scatter plots in the upper-triangular matrix follow the usual construction for correlograms. The lower-triangular matrix displays the Spearman's $\rho$'s: the mean estimate appears as the printed numeral and as the area of the solid circle; the bounds of the 95\%-confidence interval appear as the area of the internal and external dashed circles. Negative correlations appear in red. 

The correlation between different test datasets is far from perfect. That is, perhaps, obvious, but must be stressed, since it reveals that \textit{naively} hyperoptimizing a model on one test set will not necessarily generalize to other data. The relationship between splits of different datasets is more subtle. Note how the correlation between the validation and the test splits of ISIC 2017 Challenge, and the dermoscopic and clinical splits of EDRA have the highest correlations. This suggests that results measured on splits of the same dataset may not wholly generalize over data of the same type obtained on different conditions. Both phenomena show how hyperoptimizing on test gives unwarranted advantages, leading to overoptimistic assessments. \textcolor{black}{The correlogram on \figurename~\ref{fig:correlations-b} employs a different metric, the melanoma AUC, and leads to the same conclusions.}

\begin{figure*}[ht!]
    \centering
    \subfloat[Correlogram of mean melanoma/keratosis AUC across test datasets.]{\includegraphics[width=0.5\textwidth]{figures/test_dataset_correlogram_isbi.pdf}\label{fig:correlations-a}} 
     \subfloat[Correlogram of melanoma AUC across test datasets.]{\includegraphics[width=0.5\textwidth]{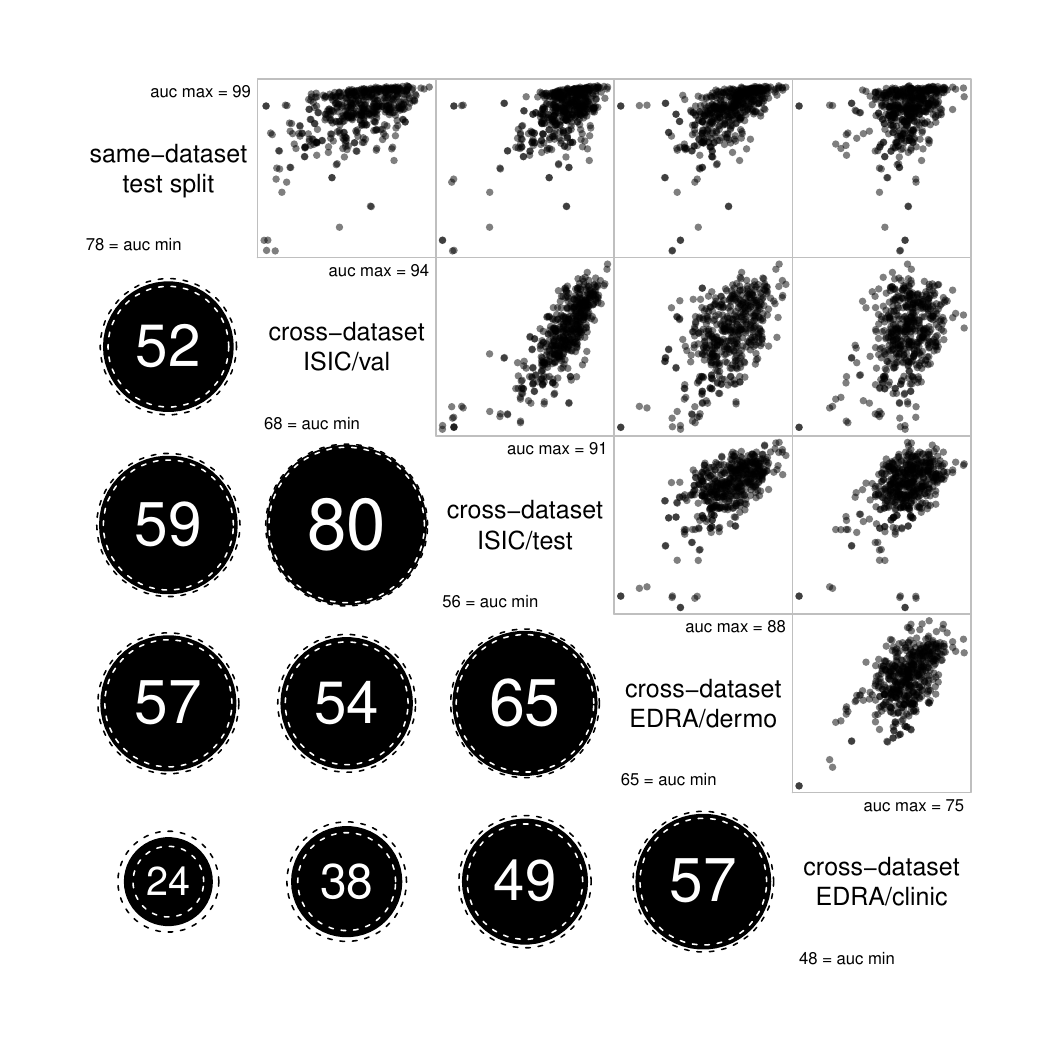}\label{fig:correlations-b}}\\ 
    \subfloat[Correlogram of metrics on ISIC 2017 Test dataset across metrics.]{\includegraphics[width=0.5\textwidth]{figures/metrics_correlogram_isic_test.pdf}\label{fig:correlations-c}} 
    \subfloat[Correlogram of metrics on EDRA Dermoscopic dataset across metrics.]{\includegraphics[width=0.5\textwidth]{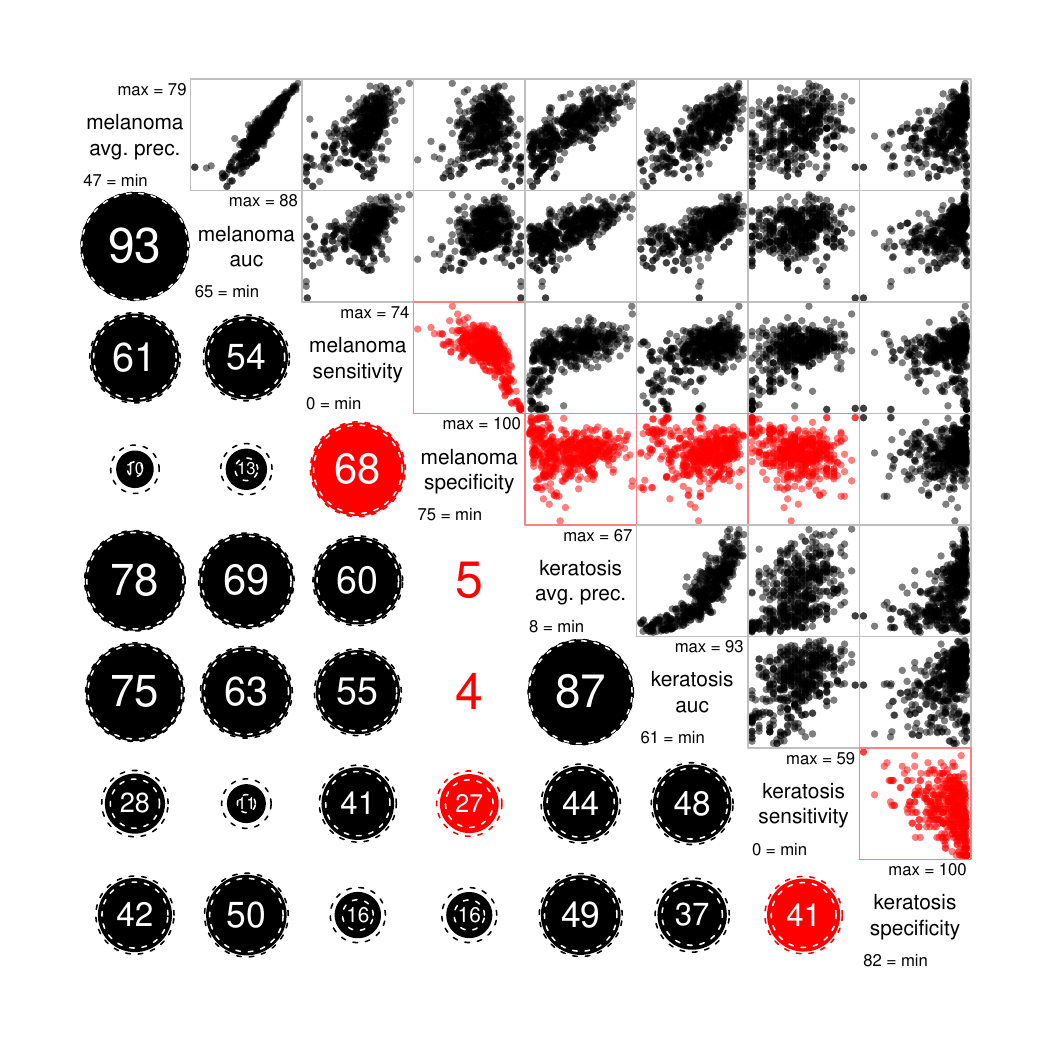}\label{fig:correlations-d}} 
    \caption{Correlograms with pair-wise correlation analyses. Sets appear on the diagonal; upper matrices show the scatter plots, and lower matrices show the Spearman correlation of each pair of sets. On lower matrices, numbers and solid circles' areas show the mean estimates, and dashed circles' areas show the 95\%-confidence bounds. Non-significant estimates appear without the circles. All numbers in \%, negative correlations are in red.}
    \label{fig:correlations}
\end{figure*}

In order to better understand the metrics, we plotted the correlogram on \figurename~\ref{fig:correlations-c}, considering only the results for the test split of the ISIC Challenge. Different metrics appear in the diagonal: average precision, area under the ROC curve, sensitivity (true positive rate), and specificity (true negative rate), for both melanoma and keratosis. The interpretation of the plots, numerals, circles, and colors is the same as above.

This correlogram is interesting for showing that many metrics have correlations that are not that big. Particularly noteworthy is the specificity, which has not only a negative correlation with sensitivity (as expected), but also a negative or very small correlation with most of the other metrics. \textcolor{black}{Such behavior is not particular to the choice of test dataset, as we illustrate in \figurename~\ref{fig:correlations-d}, which shows the same correlogram for the EDRA Dermoscopic test dataset, with similar results.}

%
%
\section{Conclusions} \label{sec:conclusion}

\textcolor{black}{In this paper, we performed an extensive investigation of the hyperparameterization of deep learning for automated melanoma screening. Our finds both validate with a rigorous experimental design previous intuitions present in existing art, and bring novel perspectives to the design of such models.}

\textcolor{black}{\textbf{The pitfalls of hyperoptimizing on test:} Although the problems with the privileged procedure become clear when we contrast it side-by-side with the blind procedure, hyperoptimizing on test is widely present machine-learning literature (we are often guilty ourselves).} It does not result from researchers' desire to cheat, but from our natural tendency to exploit scarce existing data as much as possible. Salzberg has warned researchers about the dangers of ``repeated tuning'' since 1997; his work~\cite{salzberg1997comparing} is often cited, but the issue is still far from solved.

Avoiding that pitfall requires a very strong commitment, which researchers seem unable to keep. That reinforces the importance of regular curated challenges --- like the ISIC Challenge --- in which the test set is withheld at least until the evaluations are over. The ImageNet competition is perhaps the best example of the extraordinary impact of having such a curated competition every year. 

Our findings explain, in part, why performances observed in practice fall much shorter of the numbers we get in our labs. In a single round of experiments, analyzing only two levels for nine factors, the unwarranted advantage of hyperoptimizing on test is already notable. In actual research, with many rounds of experiments over dozens of factors and hundreds of levels, the gap may be much wider. 

\textcolor{black}{\textbf{Deep-learning hyperparameterization:} in our main results, we evaluate 2560 different models, showing} that the amount of data alone explains almost half the variability in performance. That reinforces the deep-learning creed on the ``unreasonable power of data'' and has important consequences for the skin lesion analysis community: in order to move research forward, we need to curate larger shared datasets. The ISIC Archive is an essential step in that direction --- but it would have to grow almost tenfold to match the largest (private) dataset reported in the literature~\cite{esteva2017dermatologist}.

Despite the predominance of data, other factors appeared relevant. The most noteworthy is, perhaps, the use of data augmentation on test samples. The use of deeper models \textit{in combination} with extra data, also appeared as an important advantage. Increased resolution --- even in a very limited scheme --- was also advantageous. 

Notable \emph{negative} results were the use of segmentation to help classification, and the use of an extra SVM Layer. The negative result on segmentation is the most surprising, and needs further exploration, since literature reports many different ways to incorporate segmentation into classification, often with improved performances. We would like to explore that theme in future works, exhaustive in that particular scope.  

\textcolor{black}{\textbf{The importance of transfer learning:} our evaluation of transfer learning, comprising the evaluation of 1280 models, reinforce its importance, showing that the use of the technique can explain more than 60\% of the relative variation of performance.}

\textcolor{black}{\textbf{The pitfalls of sequential designs:} we demonstrate the disadvantages of the widespread procedure of sequentially optimizing the hyperparameters one by one, after picking an arbitrary starting value for them. Our results show that such procedure is very unstable, depending at random both on the starting point and the sequence of optimization.}

\textcolor{black}{\textbf{The power of ensembles:} fortunately, our experiments on the power of ensembles bring encouraging news, providing reliable performances, without neither the expense of the full factorial design, nor the instability of the sequential optimization.} Even a simple accumulation of enough randomly-sampled models was sufficient to provide adequate performance. Learning the best models on one dataset, however, helps to select the best ensembles for other datasets. Ensembles appear as a promising avenue for future explorations. In future works, we would like to design and evaluate techniques more sophisticated than pooling the decisions of the models, like model stacking, or boosting.

\section*{Acknowledgment}
We gratefully acknowledge the donation of a Tesla K40 and two TITAN X GPUs by NVIDIA Corporation, used in this work. We thank Fabio Perez, Micael Carvalho, and Fillipe D. M. de Souza, for help in revising the manuscript. We thank Prof. M. Emre Celebi for kindly providing the machine-readable metadata of the EDRA Interactive Atlas of Dermoscopy and for the help in revising the first draft. 

\section*{Funding}
E. Valle and M. Fornaciali were partially funded by Google Research Awards for Latin America 2016 \& 2017; E. Valle is also partially funded by a CNPq PQ-2 grant (311905/2017-0). A. Menegola was funded by CNPq. S. Avila is partially funded by Google Research Awards for Latin America 2018, FAPESP (2017/16246-0) and FAEPEX (3125/17). This project is partially funded by CNPq Universal grant (424958/2016-3). RECOD Lab. is partially supported by diverse projects and grants from FAPESP, CNPq, and CAPES. The funding sources had no involvement in the data acquisition, study design, result analysis, or in the manuscript writing.

\ifCLASSOPTIONcaptionsoff
  \newpage
\fi

\balance
\bibliographystyle{IEEEtran}
\bibliography{references}

\begin{thebibliography}{10}
\providecommand{\url}[1]{#1}
\csname url@samestyle\endcsname
\providecommand{\newblock}{\relax}
\providecommand{\bibinfo}[2]{#2}
\providecommand{\BIBentrySTDinterwordspacing}{\spaceskip=0pt\relax}
\providecommand{\BIBentryALTinterwordstretchfactor}{4}
\providecommand{\BIBentryALTinterwordspacing}{\spaceskip=\fontdimen2\font plus
\BIBentryALTinterwordstretchfactor\fontdimen3\font minus
  \fontdimen4\font\relax}
\providecommand{\BIBforeignlanguage}[2]{{%
\expandafter\ifx\csname l@#1\endcsname\relax
\typeout{** WARNING: IEEEtranS.bst: No hyphenation pattern has been}%
\typeout{** loaded for the language `#1'. Using the pattern for}%
\typeout{** the default language instead.}%
\else
\language=\csname l@#1\endcsname
\fi
#2}}
\providecommand{\BIBdecl}{\relax}
\BIBdecl

\bibitem{abbas2013melanoma}
Q.~Abbas, M.~Emre~Celebi, I.~F. Garcia, and W.~Ahmad, ``Melanoma recognition
  framework based on expert definition of abcd for dermoscopic images,''
  \emph{Skin Research and Technology}, vol.~19, no.~1, 2013.

\bibitem{argenziano2002dermoscopy}
G.~Argenziano, H.~P. Soyer, V.~De~Giorgi, D.~Piccolo, P.~Carli, M.~Delfino
  \emph{et~al.}, ``Dermoscopy: a tutorial,'' \emph{EDRA, Medical Publishing \&
  New Media}, 2002.

\bibitem{ballerini2013color}
L.~Ballerini, R.~B. Fisher, B.~Aldridge, and J.~Rees, ``A color and texture
  based hierarchical {K-NN} approach to the classification of non-melanoma skin
  lesions,'' in \emph{Color Medical Image Analysis}, 2013, pp. 63--86.

\bibitem{bi2017automatic}
L.~Bi, J.~Kim, E.~Ahn, and D.~Feng, ``Automatic skin lesion analysis using
  large-scale dermoscopy images and deep residual networks,'' \emph{arXiv
  preprint arXiv:1703.04197}, 2017.

\bibitem{carvalho2015}
M.~Carvalho, ``Transfer schemes for deep learning in image classification,''
  Master's thesis, University of Campinas, 2015.

\bibitem{celebi2007methodological}
M.~E. Celebi, H.~A. Kingravi, B.~Uddin, H.~Iyatomi, Y.~A. Aslandogan, W.~V.
  Stoecker, and R.~H. Moss, ``A methodological approach to the classification
  of dermoscopy images,'' \emph{Computerized Medical Imaging and Graphics},
  vol.~31, no.~6, pp. 362--373, 2007.

\bibitem{codella2015deep}
N.~Codella, J.~Cai, M.~Abedini, R.~Garnavi, A.~Halpern, and J.~R. Smith, ``Deep
  learning, sparse coding, and svm for melanoma recognition in dermoscopy
  images,'' in \emph{International Workshop on Machine Learning in Medical
  Imaging}, 2015, pp. 118--126.

\bibitem{codella2017skin}
N.~C. Codella, D.~Gutman, M.~E. Celebi, B.~Helba, M.~A. Marchetti, S.~W. Dusza,
  A.~Kalloo, K.~Liopyris, N.~Mishra, H.~Kittler \emph{et~al.}, ``{Skin Lesion
  Analysis Toward Melanoma Detection: A Challenge at the 2017 International
  Symposium on Biomedical Imaging (ISBI), Hosted by the International Skin
  Imaging Collaboration (ISIC)},'' \emph{arXiv preprint arXiv:1710.05006},
  2017.

\bibitem{codella2017deep}
N.~C. Codella, Q.-B. Nguyen, S.~Pankanti, D.~Gutman, B.~Helba, A.~Halpern, and
  J.~R. Smith, ``Deep learning ensembles for melanoma recognition in dermoscopy
  images,'' \emph{IBM Journal of Research and Development}, vol.~61, no.~4, pp.
  5:1--5:15, 2017.

\bibitem{devries2017skin}
T.~DeVries and D.~Ramachandram, ``Skin lesion classification using deep
  multi-scale convolutional neural networks,'' \emph{arXiv preprint
  arXiv:1703.01402}, 2017.

\bibitem{diaz2017incorporating}
I.~G. D{\'\i}az, ``Incorporating the knowledge of dermatologists to
  convolutional neural networks for the diagnosis of skin lesions,''
  \emph{arXiv preprint arXiv:1703.01976}, 2017.

\bibitem{esteva2017dermatologist}
A.~Esteva, B.~Kuprel, R.~A. Novoa, J.~Ko, S.~M. Swetter, H.~M. Blau, and
  S.~Thrun, ``Dermatologist-level classification of skin cancer with deep
  neural networks,'' \emph{Nature}, vol. 542, no. 7639, pp. 115--118, 2017.

\bibitem{fornaciali2014statistical}
M.~Fornaciali, S.~Avila, M.~Carvalho, and E.~Valle, ``Statistical learning
  approach for robust melanoma screening,'' in \emph{Conference on Graphics,
  Patterns and Images (SIBGRAPI)}, 2014, pp. 319--326.

\bibitem{fornaciali2016towards}
M.~Fornaciali, M.~Carvalho, F.~V. Bittencourt, S.~Avila, and E.~Valle,
  ``Towards automated melanoma screening: Proper computer vision \& reliable
  results,'' \emph{arXiv preprint arXiv:1604.04024}, 2016.

\bibitem{ge2017exploiting}
Z.~Ge, S.~Demyanov, B.~Bozorgtabar, M.~Abedini, R.~Chakravorty, A.~Bowling, and
  R.~Garnavi, ``Exploiting local and generic features for accurate skin lesions
  classification using clinical and dermoscopy imaging,'' in \emph{IEEE
  International Symposium on Biomedical Imaging (ISBI)}, 2017, pp. 986--990.

\bibitem{harangi2017skin}
B.~Harangi, ``Skin lesion detection based on an ensemble of deep convolutional
  neural network,'' \emph{arXiv preprint arXiv:1705.03360}, 2017.

\bibitem{He2016eccv}
K.~He, X.~Zhang, S.~Ren, and J.~Sun, ``Identity mappings in deep residual
  networks,'' in \emph{European Conference on Computer Vision}, 2016, pp.
  630--645.

\bibitem{iyatomi2008improved}
H.~Iyatomi, H.~Oka, M.~E. Celebi, M.~Hashimoto, M.~Hagiwara, M.~Tanaka, and
  K.~Ogawa, ``An improved internet-based melanoma screening system with
  dermatologist-like tumor area extraction algorithm,'' \emph{Computerized
  Medical Imaging and Graphics}, vol.~32, no.~7, pp. 566--579, 2008.

\bibitem{jia2017skin}
X.~Jia and L.~Shen, ``Skin lesion classification using class activation map,''
  \emph{arXiv preprint arXiv:1703.01053}, 2017.

\bibitem{kawahara2016deep}
J.~Kawahara, A.~BenTaieb, and G.~Hamarneh, ``Deep features to classify skin
  lesions,'' in \emph{IEEE International Symposium on Biomedical Imaging},
  2016, pp. 1397--1400.

\bibitem{LeCun2015deep}
Y.~LeCun, Y.~Bengio, and G.~Hinton, ``Deep learning,'' \emph{Nature}, vol. 521,
  no. 7553, pp. 436--444, 2015.

\bibitem{litjens2017survey}
G.~Litjens, T.~Kooi, B.~E. Bejnordi, A.~A.~A. Setio, F.~Ciompi, M.~Ghafoorian,
  J.~A. der Laak, B.~G. Clara, and I.~S\'anchez, ``A survey on deep learning in
  medical image analysis,'' \emph{Medical Image Analysis}, vol.~42, pp. 60--88,
  2017.

\bibitem{lopez2017skin}
A.~R. Lopez, X.~Giro-i Nieto, J.~Burdick, and O.~Marques, ``Skin lesion
  classification from dermoscopic images using deep learning techniques,'' in
  \emph{IEEE International Conference on Biomedical Engineering (BioMed)},
  2017, pp. 49--54.

\bibitem{matsunaga2017image}
K.~Matsunaga, A.~Hamada, A.~Minagawa, and H.~Koga, ``Image classification of
  melanoma, nevus and seborrheic keratosis by deep neural network ensemble,''
  \emph{arXiv preprint arXiv:1703.03108}, 2017.

\bibitem{mendoncca2013ph}
T.~Mendon{\c{c}}a, P.~M. Ferreira, J.~S. Marques, A.~R. Marcal, and J.~Rozeira,
  ``{PH2 A} dermoscopic image database for research and benchmarking,'' in
  \emph{IEEE Engineering in Medicine and Biology Society}, 2013, pp.
  5437--5440.

\bibitem{menegola2017knowledge}
A.~Menegola, M.~Fornaciali, R.~Pires, F.~V. Bittencourt, S.~Avila, and
  E.~Valle, ``Knowledge transfer for melanoma screening with deep learning,''
  in \emph{IEEE International Symposium on Biomedical Imaging}, 2017, pp.
  297--300.

\bibitem{menegola2017recod}
A.~Menegola, J.~Tavares, M.~Fornaciali, L.~T. Li, S.~Avila, and E.~Valle,
  ``{RECOD Titans at ISIC Challenge 2017},'' \emph{arXiv preprint
  arXiv:1703.04819}, 2017.

\bibitem{nasr2016melanoma}
E.~Nasr-Esfahani, S.~Samavi, N.~Karimi, S.~M.~R. Soroushmehr, M.~H. Jafari,
  K.~Ward, and K.~Najarian, ``Melanoma detection by analysis of clinical images
  using convolutional neural network,'' in \emph{IEEE Engineering in Medicine
  and Biology Society}, 2016, pp. 1373--1376.

\bibitem{perez2018data}
F.~Perez, C.~Vasconcelos, S.~Avila, and E.~Valle, ``Data augmentation for skin
  lesion analysis,'' in \emph{{ISIC Skin Image Analysis Workshop}}, 2018.

\bibitem{RonnebergerFB15}
O.~Ronneberger, P.~Fischer, and T.~Brox, ``{U-Net}: Convolutional networks for
  biomedical image segmentation,'' in \emph{Medical Image Computing and
  Computer Assisted Intervention}, 2015, pp. 234--241.

\bibitem{sabbaghi2016deep}
S.~Sabbaghi, M.~Aldeen, and R.~Garnavi, ``A deep bag-of-features model for the
  classification of melanomas in dermoscopy images,'' in \emph{IEEE Engineering
  in Medicine and Biology Society}, 2016, pp. 1369--1372.

\bibitem{salzberg1997comparing}
S.~L. Salzberg, ``On comparing classifiers: Pitfalls to avoid and a recommended
  approach,'' \emph{Data Mining and Knowledge Discovery}, vol.~1, no.~3, pp.
  317--328, 1997.

\bibitem{szegedy2016inceptionv4}
C.~Szegedy, S.~Ioffe, V.~Vanhoucke, and A.~A. Alemi, ``Inception-v4,
  inception-resnet and the impact of residual connections on learning,''
  \emph{AAAI Conference on Artificial Intelligence}, pp. 4278--4284, 2017.

\bibitem{torralba2011unbiased}
A.~Torralba and A.~A. Efros, ``Unbiased look at dataset bias,'' in \emph{IEEE
  Conference on Computer Vision and Pattern Recognition}, 2011, pp. 1521--1528.

\bibitem{vasconcelos2017increasing}
C.~N. Vasconcelos and B.~N. Vasconcelos, ``Increasing deep learning melanoma
  classification by classical and expert knowledge based image transforms,''
  \emph{arXiv preprint arXiv:1702.07025}, 2017.

\bibitem{wighton2011generalizing}
P.~Wighton, T.~K. Lee, H.~Lui, D.~I. McLean, and M.~S. Atkins, ``Generalizing
  common tasks in automated skin lesion diagnosis,'' \emph{IEEE Transactions on
  Information Technology in Biomedicine}, vol.~15, no.~4, pp. 622--629, 2011.

\bibitem{yang2017novel}
X.~Yang, Z.~Zeng, S.~Y. Yeo, C.~Tan, H.~L. Tey, and Y.~Su, ``A novel multi-task
  deep learning model for skin lesion segmentation and classification,''
  \emph{arXiv preprint arXiv:1703.01025}, 2017.

\bibitem{yoshida2016simple}
T.~Yoshida, M.~E. Celebi, G.~Schaefer, and H.~Iyatomi, ``Simple and effective
  pre-processing for automated melanoma discrimination based on cytological
  findings,'' in \emph{IEEE International Conference on Big Data}, 2016, pp.
  3439--3442.

\bibitem{yu2017automated}
L.~Yu, H.~Chen, Q.~Dou, J.~Qin, and P.-A. Heng, ``Automated melanoma
  recognition in dermoscopy images via very deep residual networks,''
  \emph{IEEE Transactions on Medical Imaging}, vol.~36, no.~4, pp. 994--1004,
  2017.

\end{thebibliography}

\end{document}